\documentclass[sigconf]{acmart}

\usepackage{booktabs} 

\usepackage{amssymb,amsmath}
\usepackage{algorithm}  
\usepackage{algpseudocode}   
\usepackage{caption}
\usepackage{subfigure}
\usepackage{textcomp}

\setcopyright{rightsretained}

\acmDOI{10.475/123_4}

\acmISBN{123-4567-24-567/08/06}

\acmConference[ACM Sample'2018]{ACM Woodstock conference}
\acmYear{2018}
\copyrightyear{2018}

\acmArticle{4}
\acmPrice{15.00}

\begin{document}
\title{Deep Reinforcement Learning for Sponsored Search Real-time Bidding}

\author{Jun Zhao}
\affiliation{\institution{Alibaba Group}}

\author{Guang Qiu}
\affiliation{\institution{Alibaba Group}}

\author{Ziyu Guan}
\affiliation{\institution{Xidian University}}

\author{Wei Zhao}
\affiliation{\institution{Xidian University}}

\author{Xiaofei He}
\affiliation{\institution{Zhejiang University}}

\renewcommand{\shortauthors}{J. Zhao et al.}

\begin{abstract}
Bidding optimization is one of the most critical problems in online advertising. Sponsored search (SS) auction, due to the randomness of user query behavior and platform nature, usually adopts keyword-level bidding strategies. In contrast, the display advertising (DA), as a relatively simpler scenario for auction, has taken advantage of real-time bidding (RTB) to boost the performance for advertisers. In this paper, we consider the RTB problem in sponsored search auction, named SS-RTB. SS-RTB has a much more complex dynamic environment, due to stochastic user query behavior and more complex bidding policies based on multiple keywords of an ad. Most previous methods for DA cannot be applied. We propose a reinforcement learning (RL) solution for handling the complex dynamic environment. Although some RL methods have been proposed for online advertising, they all fail to address the ``environment changing'' problem: the state transition probabilities vary between two days. Motivated by the observation that auction sequences of two days share similar transition patterns at a proper aggregation level, we formulate a robust MDP model at hour-aggregation level of the auction data and propose a control-by-model framework for SS-RTB. Rather than generating bid prices directly, we decide a bidding model for impressions of each hour and perform real-time bidding accordingly. We also extend the met- hod to handle the multi-agent problem. We deployed the SS-RTB system in the e-commerce search auction platform of Alibaba. Empirical experiments of offline evaluation and online A/B test demonstrate the effectiveness of our method.
\end{abstract}

%
%



\keywords{Bidding Optimization, Sponsored Search, Reinforcement Learning}

\maketitle


\section{Introduction}

Bidding optimization is one of the most critical problems for maximizing advertisers' profit in online advertising. In the sponsored search (SS) scenario, the problem is typically formulated as an optimization of the advertisers' objectives (KPIs) via seeking the best settings of keyword bids \cite{borgs2007dynamics,feldman2007budget}. The keyword bids are usually assumed to be fixed during the online auction process. However, the sequence of user queries (incurring \emph{impressions} and \emph{auctions} for online advertising) creates a complicated dynamic environment where a real-time bidding strategy could significantly boost advertisers' profit. This is more important on e-commerce auction platforms since impressions more readily turn into purchases, compared to traditional web search. In this work, we consider the problem, Sponsored Search Real-Time Bidding (SS-RTB), which aims to generate proper bids at the impression level in the context of SS. To the best of our knowledge, there is no publicly available solution to SS-RTB. 



The RTB problem has been studied in the context of display advertising (DA). Nevertheless, SS-RTB is intrinsically different from RTB. In DA, the impressions for bidding are concerned with ad placements in publishers' web pages, while in SS the targets are ranking lists of dynamic user queries. The key difference are: (1) for a DA impression only the winning ad can be presented to the user (i.e. a 0-1 problem), while in the SS context multiple ads which are ranked high can be exhibited to the query user; (2) In SS, we need to adjust bid prices on multiple keywords for an ad to achieve optimal performance, while an ad in DA does not need to consider such a keyword set. These differences render popular methods for RTB in DA, such as predicting winning market price \cite{wu2015predicting} or winning rate \cite{zhang2014optimal}, inapplicable in SS-RTB. Moreover, compared to ad placements in web pages, user query sequences in SS are stochastic and highly dynamic in nature. This calls for a complex model for SS-RTB, rather than the shallow models often used in RTB for DA \cite{zhang2014optimal}. 


One straightforward solution for SS-RTB is to establish an optimization problem that outputs the optimal bidding setting for each impression independently. However, each impression bid is strategically correlated by several factors given an ad, including the ad's budget and overall profit, and the dynamics of the underlying environment. The above greedy strategy often does not lead to a good overall profit \cite{cai2017real}. Thus, it is better to model the bidding strategy as a sequential decision on the sequence of impressions in order to optimize the overall profit for an ad by considering the factors mentioned above. This is exactly what reinforcement learning (RL) \cite{mnih2013playing,poole2010artificial} does. By RL, we can model the bidding strategy as a dynamic interactive control process in a complex environment rather than an independent prediction or optimization process. The budget of an ad can be dynamically allocated across the sequence of impressions, so that both immediate auction gain and long-term future rewards are considered.



\begin{figure} 
\includegraphics[height=2in, width=3.2in]{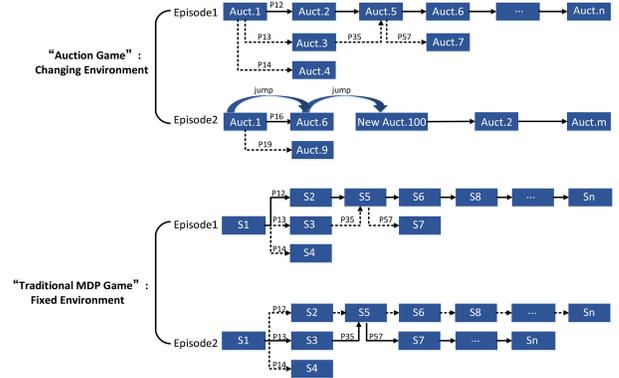}
\caption{Auction Environment vs Game Environment}
\label{envp}
\end{figure}

Researchers have explored using RL in online advertising. Amin \emph{et al.} constructed a Markov Decision Process (MDP) for budget optimization in SS \cite{amin2012budget}. Their method deals with impressions/auctions in a batch model and hence cannot be used for RTB. Moreover, the underlying environment for MDP is ``static'' in that all states share the same set of transition probabilities and they do not consider impression-specific features. Such a MDP cannot well capture the complex dynamics of auction sequences in SS, which is important for SS-RTB. Cai \emph{et al.} developed a RL method for RTB in DA \cite{cai2017real}. The method combines an optimized reward for the current impression (based on impression-level features) and the estimate of future rewards by a MDP for guiding the bidding process. The MDP is still a static one as in \cite{amin2012budget}. Recently, a deep reinforcement learning (DRL) method was proposed in \cite{wang2017ladder} for RTB in DA. Different from the previous two works, their MDP tries to fully model the dynamics of auction sequences. However, they also identified an ``environment changing'' issue: the underlying dynamics of the auction sequences from two days could be very different. For example, the auction number and the users' visits could heavily deviate between days. A toy illustration is presented in Figure~\ref{envp}. Compared to a game environment, the auction environment in SS is itself stochastic due to the stochastic behaviors of users. Hence, the model learned from Day 1 cannot well handle the data from Day 2. Although \cite{wang2017ladder} proposed a sampling mechanism for this issue, it is still difficult to guarantee we obtain the same environment for different days. Another challenge that existing methods fail to address is the multi-agent problem. That is, there are usually many ads competing with one another in auctions of SS. Therefore, it is important to consider them jointly to achieve better global bidding performance.

Motivated by the challenges above, this paper proposes a new DRL method for the SS-RTB problem. In our work, we captured various discriminative information in impressions such as market price and conversion rate (CVR), and also try to fully capture the dynamics of the underlying environment. The core novelty of our proposed method lies in how the environment changing problem is handled. We solve this problem by observing the fact that statistical features of proper aggregation (e.g. by hour) of impressions has strong periodic state transition patterns in contrast to impression level data. Inspired by this, we design a robust MDP at the hour-aggregation level to represent the sequential decision process in the SS auction environment. At each state of the MDP, rather than generating bid prices directly, we decide a bidding model for impressions of that hour and perform real-time bidding accordingly. In other words, the robust MDP aims to learn the optimal parameter policy to control the real-time bidding model. Different from the traditional ``control-by-action'' paradigm of RL, we call this scheme ``control-by-model''. By this control-by-model learning scheme, our system can do real-time bidding via capturing impression-level features, and meanwhile also take the advantage of RL to periodically control the bidding model according to the real feedback from the environment. Besides, considering there are usually a considerable number of ads, we also design a massive-agent learning algorithm by combining competitive reward and cooperative reward together.


The contribution of this work is summarized as follows: (1) We propose a novel research problem, Sponsored Search Real-Time Bidding (SS-RTB), and properly motivate it. (2) A novel deep reinforcement learning (DRL) method is developed for SS-RTB which can well handle the environment changing problem. It is worth to note that, the robust MDP we proposed is also a general idea that can be applied to other applications. We also designed an algorithm for handling the massive-agent scenario. (3) We deploy the DRL model in the Alibaba search auction platform, one of the largest e-commerce search auction platforms in China, to carry out evaluation. The offline evaluation and standard online A/B test demonstrate the superiority of our model.

\section{Related Work}
In this section, we briefly review two fields related to our work: reinforcement learning and bidding optimization.

\subsection{Reinforcement Learning}
In reinforcement learning (RL) theory, a control system is formulated as a Markov Decision Process (MDP). A MDP can be mathematically represented as a tuple <$\mathcal{S}, \mathcal{A}, p, r$>, where $\mathcal{S}$ and $\mathcal{A}$ represent the state and action space respectively, $p(\cdot)$ denotes the transition probability function, and $r(\cdot)$ denotes the feedback reward function. The transition probability from state $s \in \mathcal{S}$ to $s' \in \mathcal{S}$ by taking action $a \in \mathcal{A}$ is $p(s, a, s')$. The reward received after taking action $a$ in state $s$ is $r(s,a)$. The goal of the model is to learn an optimal policy (a sequence of decisions mapping state $s$ to action $a$), so as to maximize the expected accumulated long term reward.

Remarkably, deep neural networks coupled with RL have achieved notable success in diverse challenging tasks: learning policies to play games \cite{silver2016mastering,mnih2013playing,mnih2015human}, continuous control of robots and autonomous vehicles \cite{levine2014learning,gu2016continuous,hafner2011reinforcement}, and recently, online advertising \cite{cai2017real,wang2017ladder,amin2012budget}. While the majority of RL research has a consistent environment, applying it to online advertising is not a trivial task since we have to deal with the environment changing problem mentioned previously. The core novelty of our work is that we propose a solution which can well handle this problem.

Moreover, when two or more agents share an environment, the performance of RL is less understood. Theoretical proofs or guarantees for multi-agent RL are scarce and only restricted to specific types of small tasks \cite{schwartz2014multi,busoniu2008comprehensive,tampuu2017multiagent}. The authors in \cite{tampuu2017multiagent} investigated how two agents controlled by independent Deep Q-Networks (DQN) interact with each other in the game of Pong. They used the environment as the sole source of interaction between agents. In their study, by changing reward schema from competition to cooperation, the agents would learn to behave accordingly. In this paper, we adopt a similar idea to solve the multi-agent problem. However, the scenario in online advertising is quite different. Not only the agent number is much larger, but also the market environment is much more complex. No previous work has explored using cooperative rewards to address the multi-agent problem in such scenarios.

\subsection{Bidding optimization}
In sponsored search (SS) auctions, bidding optimization has been well studied. However, most previous works focused on the keyword-level auction paradigm, which is concerned with (but not limited to) budget allocation \cite{borgs2005multi,feldman2007budget,muthukrishnan2007stochastic}, bid generation for advanced match \cite{even2009bid,broder2011bid,fuxman2008using}, keywords' utility estimation \cite{borgs2007dynamics,kitts2004optimal}.

Unlike in SS, RTB has been a leading research topic in display advertising (DA) \cite{yuan2013real,wang2015real}. Different strategies of RTB have been proposed by researchers \cite{zhang2014optimal,wu2015predicting,cai2017real,chen2011real,lee2013real}. In \cite{zhang2014optimal}, the authors proposed a functional optimization framework to learn the optimal bidding strategy. However, their model is based on an assumption that the auction winning function has a consistent concave shape form. Wu \emph{et al.} proposed a fixed model with censored data to predict the market price in real-time \cite{wu2015predicting}. Although these works have shown significant advantage of RTB in DA, they are not applicable to the SS context due to the differences between SS and DA discussed in Section~1.

In addition to these prior studies, recently, a number of research efforts in applying RL to bidding optimization have been made \cite{amin2012budget,wang2017ladder,cai2017real}. In \cite{amin2012budget}, Amin \emph{et al.} combined the MDP formulation with the Kaplan-Meier estimator to learn the optimal bidding policy, where decisions were made on keyword level. In \cite{cai2017real}, Cai \emph{et al.} formulated the bidding decision process as a similar MDP problem, but taking one step further, they proposed a RTB method for DA by employing the MDP as a long term reward estimator. However, both of the two works considered the transition probabilities as static and failed to capture impression-level features in their MDPs. More recently, the authors in \cite{wang2017ladder} proposed an end-to-end DRL method with impression-level features formulated in the states and tried to capture the underlying dynamics of the auction environment. They used random sampling to address the environment changing problem. Nevertheless, random sampling still cannot guarantee an invariant underlying environment. Our proposed method is similar to that of \cite{wang2017ladder} in that we use a similar DQN model and also exploit impression-level features. However, the fundamental difference is that we propose a robust MDP based on hour-aggregation of impressions and a novel control-by-model learning scheme. Besides, we also try to address the multi-agent problem in the context of online advertising, which has not been done before.

\section{Problem Definition}\label{section:problem}
In this section, we will mathematically formulate the problem of real-time bidding optimization in sponsored search auction platforms (SS-RTB).

In a simple and general scenario, an ad has a set of keyword tuples $\{kwinf_1$, $kwinf_2$, $\dots$, $kwinf_m\}$, where each tuple $kwinf\_i$ can be defined as <$belong\_ad, keyword, bidprice$>. Typically, the $bidprice$ here is preset by the advertiser. The process of an auction $auct$ could then be depicted as: everytime a user $u$ visits and types a query, the platform will retrieve a list of relevant keyword tuples, [$kwinf_1$, $kwinf_2$,  $\dots$, $kwinf_r$]\footnote{Typically, an ad can only have one keyword tuple (i.e. the most relevant one) in the list}, from the ad repository for auction. Each involved ad is then assigned a ranking score according to its retrieved keyword tuple as $bidscore * bidprice$. Here, $bidscore$ is obtained from factors such as relevance and personalization, etc. Finally, top ads will be presented to the user $u$.

For SS-RTB, the key problem is to find another $opt\_bidprice$ rather than $bidprice$ for the matched keyword tuples during real-time auction, so as to maximize an ad's overall profit. Since we carry out the research in an e-commerce search auction platform, in the following we will use concepts related to the e-commerce search scenario. Nevertheless, the method is general and could be adapted to other search scenarios. We define an ad's goal as maximizing the purchase amount $PUR\_AMT_d$ as income in a day $d$, while minimizing the cost $COST_d$ as expense in $d$, with a constraint that the $PUR\_AMT_d$ should not be smaller than the advertiser's expected value $g$. We can formulate the problem as:
\begin{equation}  \label{eq:obj_1}
\begin{aligned} 
  & \max \quad PUR\_AMT_d / COST_d \\
   &     \quad    s.t.  \quad PUR\_AMT_d >= g
\end{aligned}
\end{equation}
Observing that $PUR\_AMT_d$ has highly positive correlation with $COST_d$, we can change it to:
\begin{equation}  \label{eq:obj_2}
\begin{aligned} 
      & \max \quad PUR\_AMT_d  \\
        & \quad  s.t.  \quad COST_d = c
\end{aligned}
\end{equation}
Eq.~(\ref{eq:obj_2}) is equivalent to Eq.~(\ref{eq:obj_1}) when $COST_d$ is positively correlated with $PUR\_AMT_d$ (as is the usual case). We omit the proof due to space limitation. The problem we study in this paper is to decide $opt\_bidprice$ in real-time for an ad in terms of objective (\ref{eq:obj_2}).

\section{Methodology}
\subsection{A Sketch Model of Reinforcement Learning}\label{subsection:sketch}

Based on the problem we defined in section \ref{section:problem}, we now formulate it into a sketch model of RL:

\noindent \textbf{State $s$}: We design a general representation for states as $s$ = <$b$, $t$, $\overrightarrow{auct}$>, where $b$ denotes the budget left for the ad, $t$ denotes the step number of the decision sequence, and $\overrightarrow{auct}$ is the auction (impression) related feature vector that we can get from the advertising environment. It is worth to note that, for generalization purpose, the $b$ here is not the budget preset by the advertiser. Instead, it refers to the cost that the ad expects to expend in the left steps of auctions.

\noindent \textbf{Action $a$}: The decision of generating the real-time bidding price for each auction.

\noindent \textbf{Reward $r(s, a)$}: The income (in terms of $PUR\_AMT$) gained according to a specific action $a$ under state $s$.

\noindent \textbf{Episode $ep$}: In this paper, we always treat one day as an episode.

Finally, our goal is to find a policy $\pi(s)$ which maps each state $s$ to an action $a$, to obtain the maximum expected accumulated rewards: $\sum_{i=1}^n \gamma^{i-1} r(s_i, a_i) $. $\{\gamma^i\}$ is the set of discount coefficients used in a standard RL model \cite{sutton1998reinforcement}.




Due to the randomness of user query behavior, one might never see two auctions with exactly the same feature vector. Hence, in previous work of online advertising, there is a fundamental assumption for MDP models: two auctions with similar features can be viewed as the same \cite{amin2012budget,wang2017ladder,cai2017real}. Here we provide a mathematical form of this assumption as follows.

\newtheorem{assumption}{Assumption}
\begin{assumption} \label{assumption: eq_pv}
Two auction $\overrightarrow{auct}_i$ and $\overrightarrow{auct}_j$ can be a substitute to each other as they were the same if and only if they meet the following condition:
\begin{displaymath}
\frac {\Arrowvert \overrightarrow {auct}_i - \overrightarrow {auct}_j\Arrowvert^{2}  } {\min(\Arrowvert\overrightarrow {auct}_i\Arrowvert^2,\Arrowvert\overrightarrow {auct}_j\Arrowvert^2)} < 0.01
\end{displaymath}
This kind of substitution will not affect the performance of a MDP-based control system.
\end{assumption}

However, the above sketch model is defined on the auction level. This cannot handle the environment change problem discussed previously (Figure~\ref{envp}). In other words, given day 1 and day 2, the model trained on $ep_1$ cannot be applied to $ep_2$ since the underlying environment changes. In the next, we present our solution to this problem.


\subsection{The Robust MDP model}

Our solution is inspired by a series of regular patterns observed in real-data. We found that, by viewing the sequences of auctions at an aggregation level, the underlying environments of two different days share very similar dynamic patterns.

\begin{center}
\begin{figure}
\subfigure[]
{ 
\begin{minipage}[t]{0.45\linewidth}
\includegraphics[height=25mm, width=35mm]{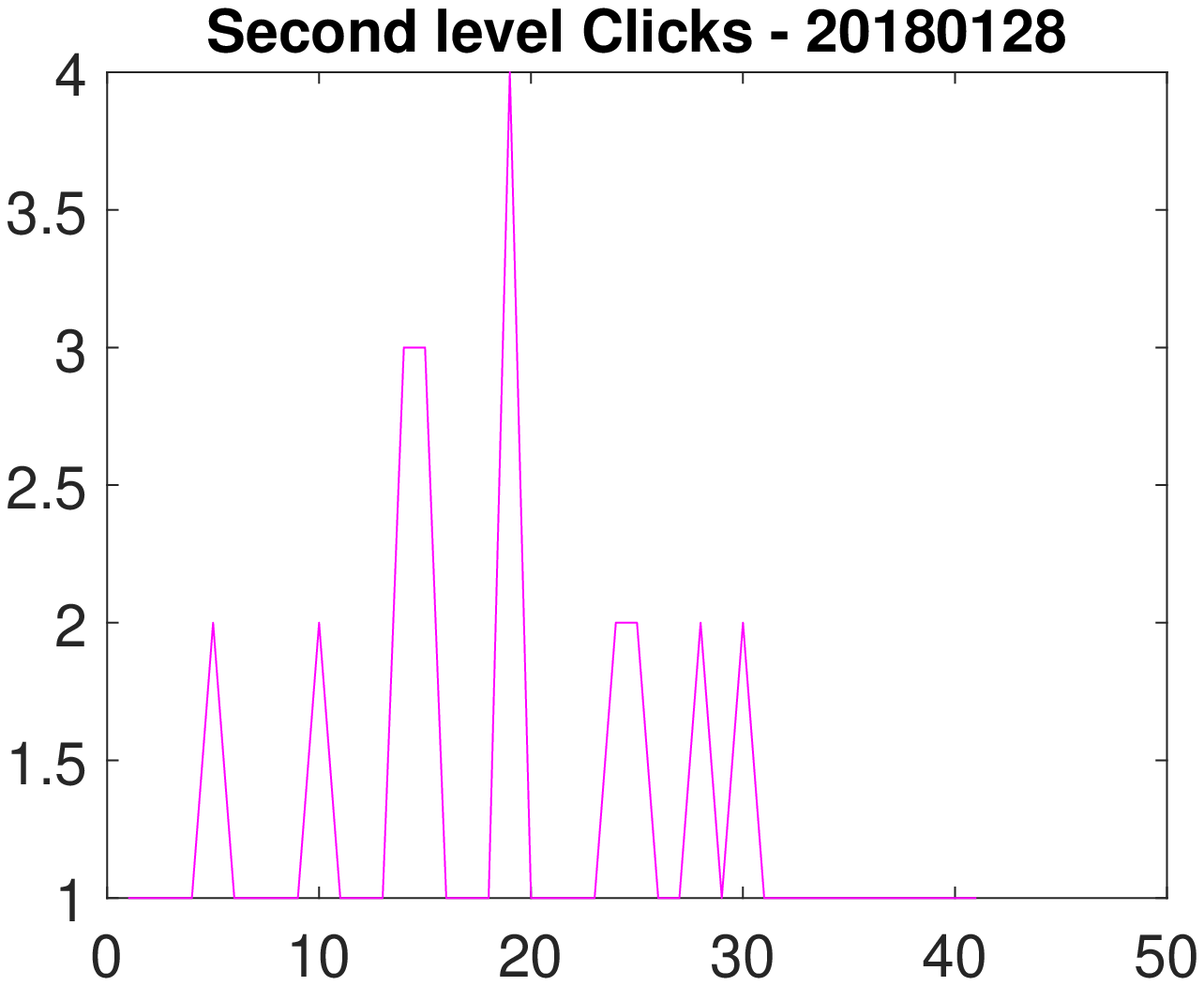}
\end{minipage}
}
\subfigure[]
{ 
\begin{minipage}[t]{0.45 \linewidth}
\includegraphics[height=25mm, width=35mm]{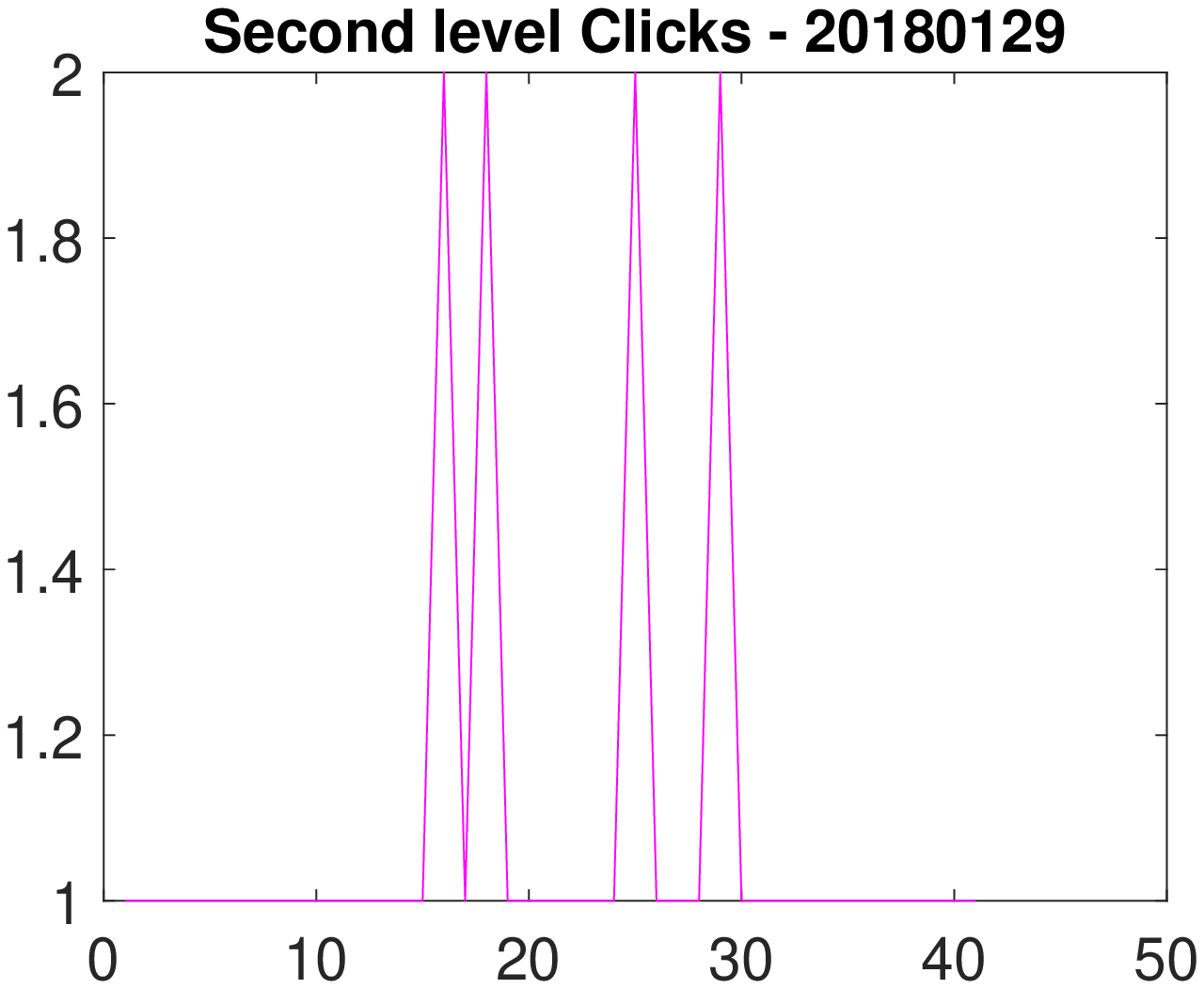}
\end{minipage}
}
\subfigure[]
{ 
\begin{minipage}[t]{0.45\linewidth}
\includegraphics[height=25mm, width=35mm]{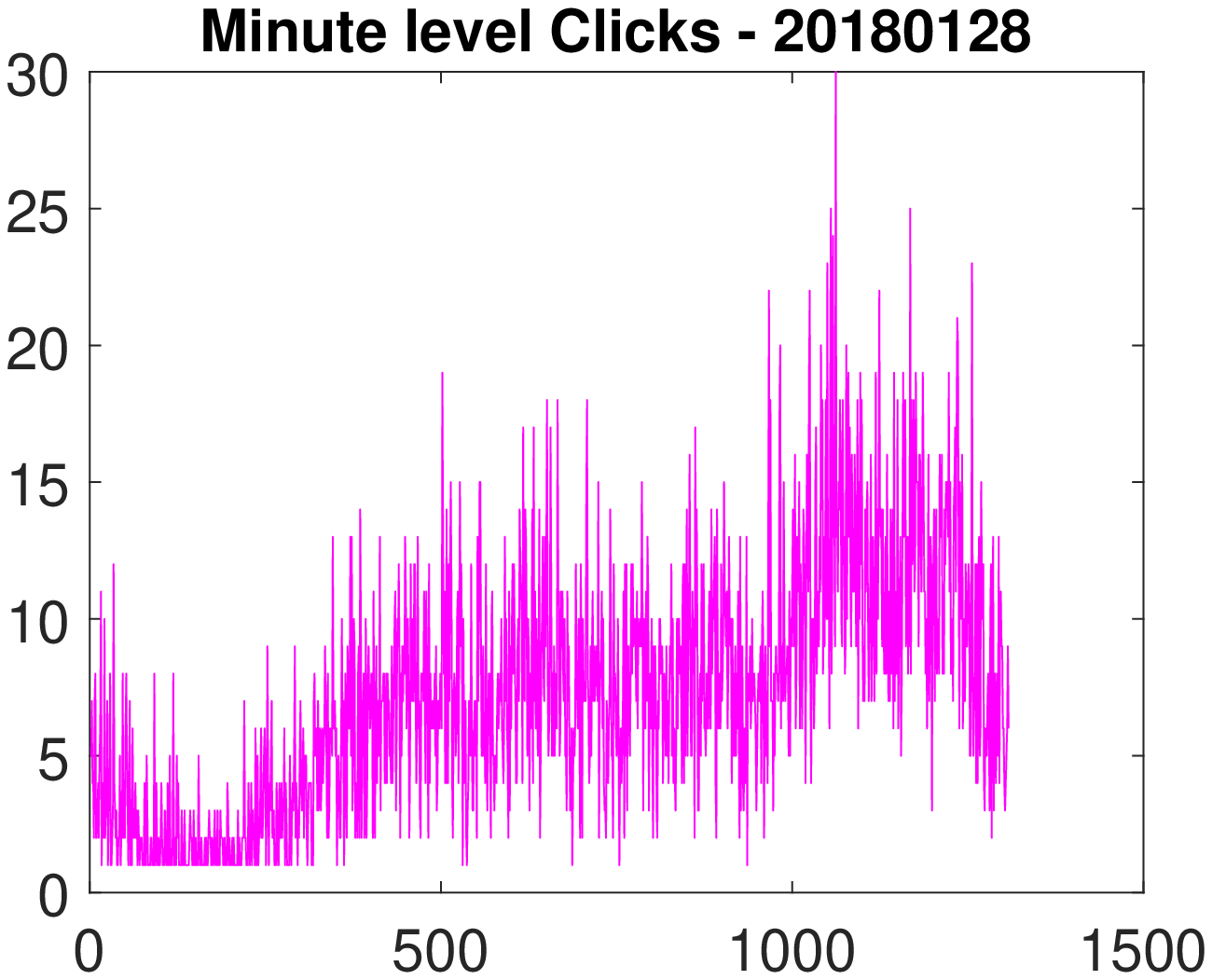}
\end{minipage}
}
\subfigure[]
{ 
\begin{minipage}[t]{0.45\linewidth}
\includegraphics[height=25mm, width=35mm]{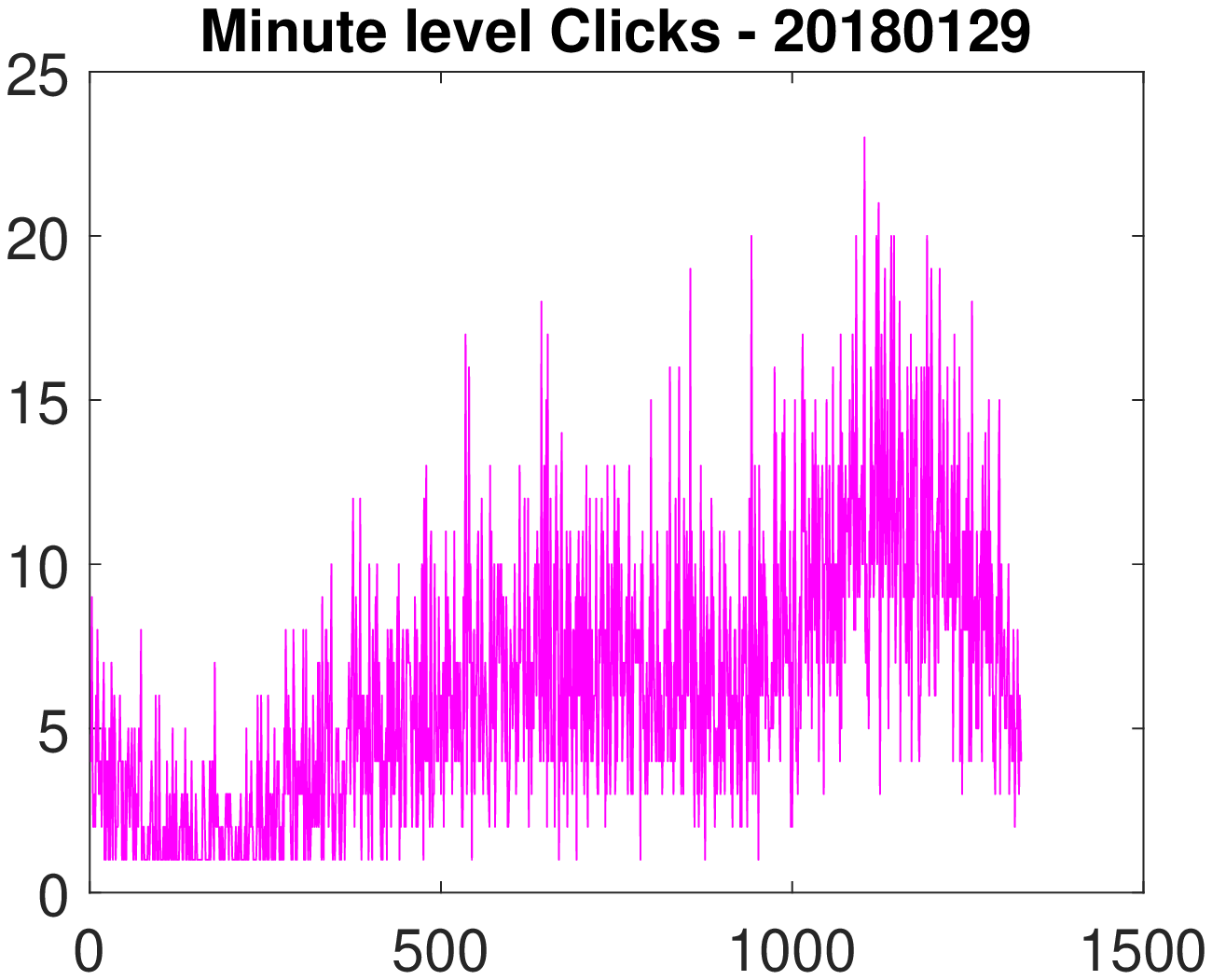}
\end{minipage}
}
\subfigure[]
{ 
\begin{minipage}[t]{0.45\linewidth}
\includegraphics[height=25mm, width=35mm]{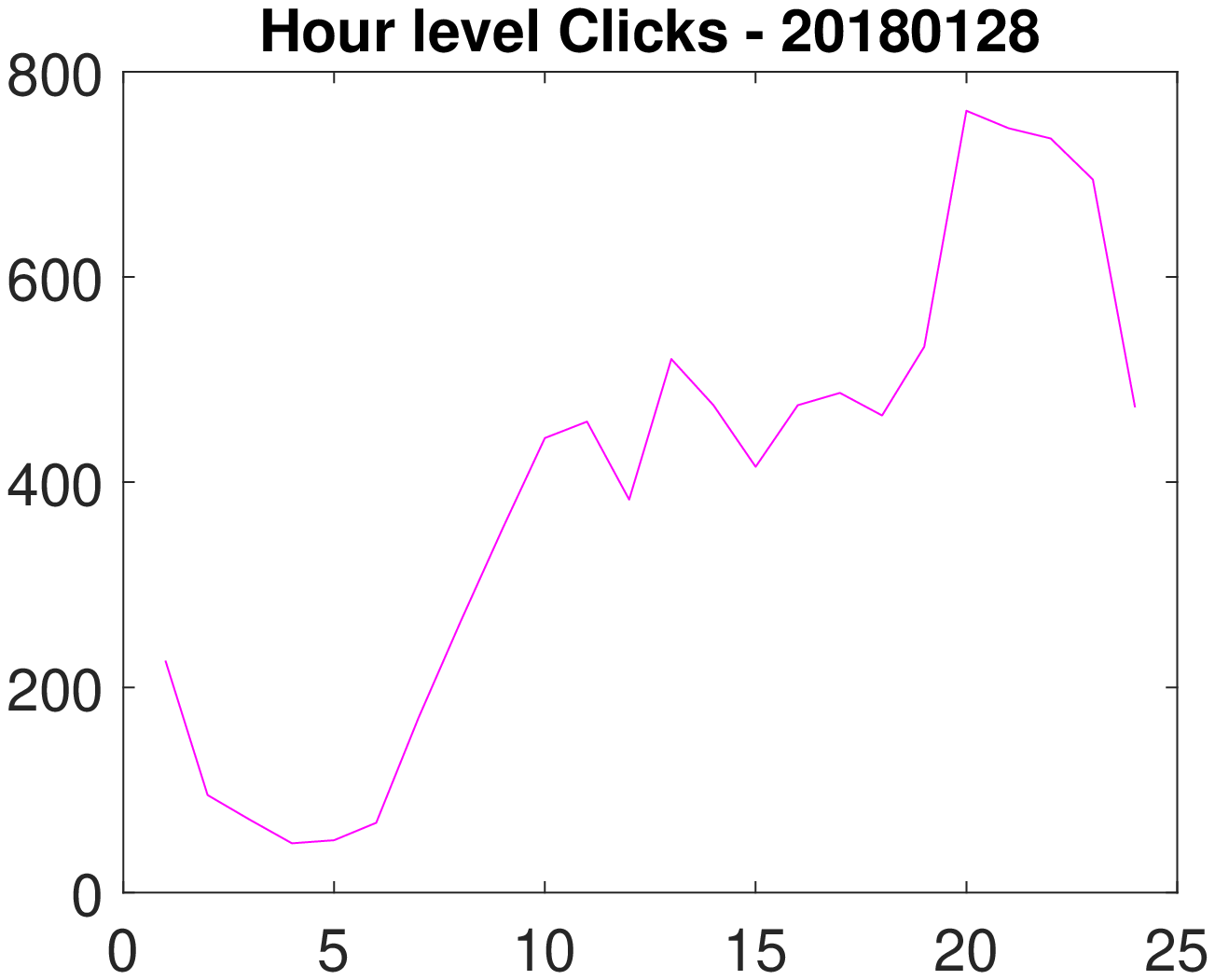}
\end{minipage}
}
\subfigure[]
{ 
\begin{minipage}[t]{0.45\linewidth}
\includegraphics[height=25mm, width=35mm]{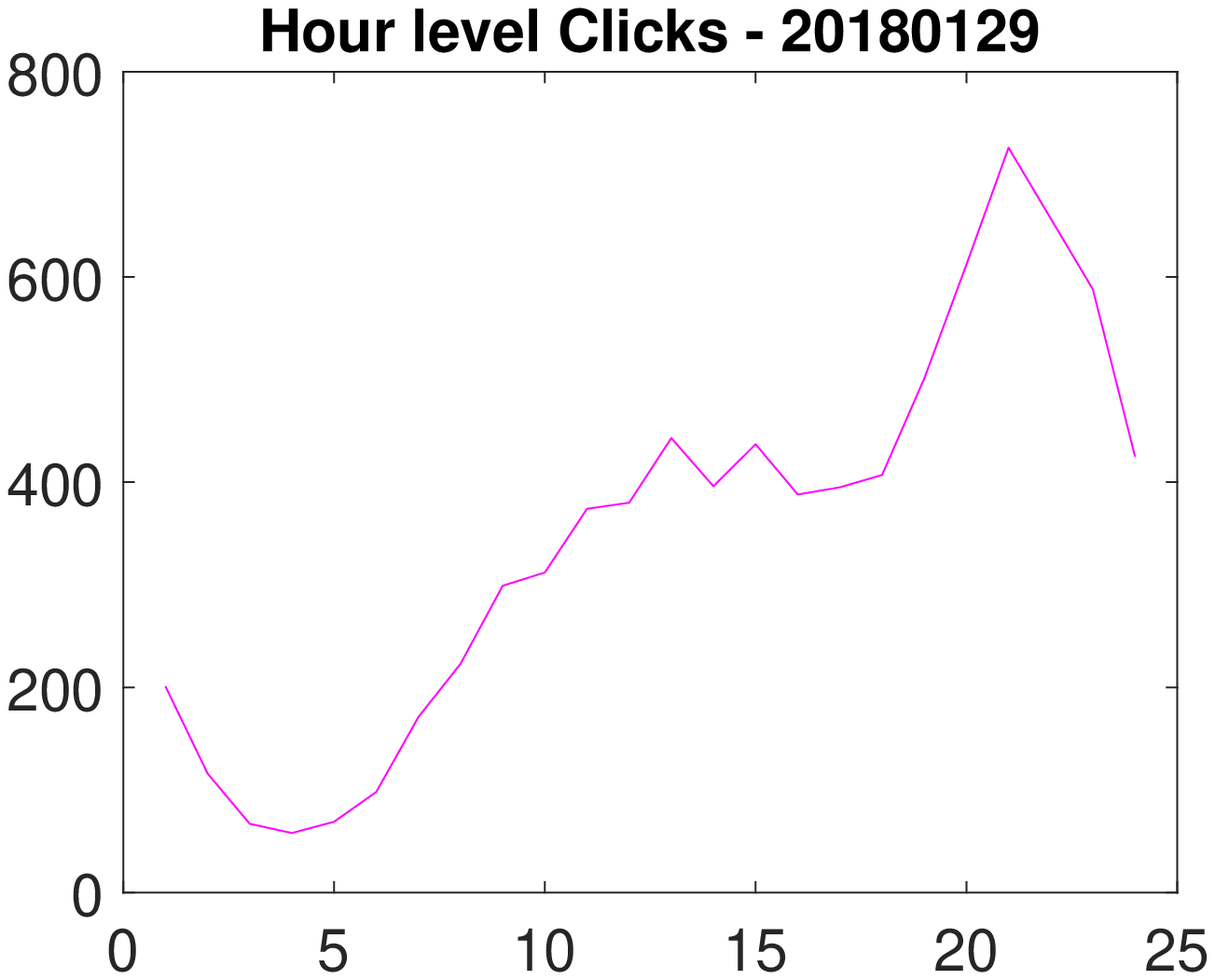}
\end{minipage}
}
\caption{Patterns of Different Level Clicks}
\label{fig:level}
\end{figure}
\end{center}

We illustrate a typical example in Figure~\ref{fig:level}, which depicts the number of clicks of an ad at different levels of aggregation (from second-level to hour-level) in Jan. 28th, 2018 and Jan. 29th, 2018 respectively. It can be observed that, the second-level curves does not exhibit a similar pattern (Figure~\ref{fig:level}(a) and (b)), while from both minute-level and hour-level we can observe a similar wave shape. In addition, it also suggests that the hour-level curves are more similar than those of minute-level. We have similar observations on other aggregated measures.

The evolving patterns of the same measure are very similar between two days at hour-level, indicating the underlying rule of this auction game is the same. Inspired by these regular patterns, we will take advantage of hour-aggregated features rather than auction-level features to formulate the MDP model. Intuitively, if we treat each day as a 24 steps of auction game, an episode of any day would always have the same general experiences. For example, it will meet an auction valley between 3:00AM to 7:00AM, while facing a heavy competitive auction peak at around 9:00AM and a huge amount of user purchases at around 8:00PM. We override the sketch MDP model as follows.

\noindent \textbf{State Transition.}
The auctions of an episode will be grouped into $m$ ($m=24$ in our case) groups according to the timestamp. Each group contains a batch of auctions in the corresponding period. A state $s$ is re-defined as <$b, t, \vec g$>, where $b$ is the left budget, $t$ is the specific time period, $\vec g$ denotes the feature vector containing aggregated statistical features of auctions in time period $t$, e.g. number of click, number of impression, cost, click-through-rate (CTR), conversion rate (CVR), pay per click (PPC), etc. In this way, we obtain an episode with fixed steps $m$. In the following, we show that the state transition probabilities are consistent between two days.

Suppose given state $s$ and action $a$, we observe the next state $s'$. We rewrite the state transition probability function $p(s, a, s')$ as
\begin{equation}  \label{eq:transition_1}
\begin{aligned} 
p(s, a, s') & = p(S = s' | S = s, A = a)  \\
                    & = p(S = (b-\delta, t+1, {\vec g}') | S = (b, t, \vec g), A = a) \\
                    & = p(T = t+1 | T = t, A = a) \\
                    &    \quad \cdot p(B = b-\delta, G = {\vec g}' | B = b, G = \vec g, A = a) \\
                    & = p(B = b-\delta, G = {\vec g}' | B = b, G = \vec g, A = a) \\
\end{aligned}
\end{equation}
where upper case letters represent random variables and $\delta$ is the cost of the step corresponding to the action $a$. Since $B$ is only affected by $\delta$ which only depends on the action $a$, Eq.~(\ref{eq:transition_1}) could be rewritten as: 
\begin{equation}  \label{eq:transition_2}
\begin{aligned} 
p(s, a, s') & = p(Cost = \delta, G = {\vec g}' | G = \vec g, A = a) \\
\end{aligned}
\end{equation}
Because $Cost$ is also designed as a feature of ${\vec g}'$, Eq.~(\ref{eq:transition_2}) then becomes:
\begin{equation}  \label{eq:transition_3}
\begin{aligned} 
p(s, a, s') & = p(G = {\vec g}' | G = \vec g, A = a) \\
                    & = \prod_i p(g_i' | g_i, A = a)
\end{aligned}
\end{equation}
where each $g_i$ represents an aggregated statistical feature and we use the property that features are independent with one another. By inspecting auctions from adjacent days, we have the following empirical observation.

\newtheorem{observation}{Observation}
\begin{observation} \label{assumption: fea}
Let $g_{i,t}$ be the aggregated value of feature $i$ at step $t$. When $g_{i,t}$ and the action $a_t$ are fixed, $g_{i,t+1}$ will meet:
\begin{displaymath}
 g_{i,t+1}  \in [(1-\eta) \bar{g},  (1+\eta) \bar{g}]
\end{displaymath}
Where $\bar{g}$ is the sample mean of $g_{i,t+1}$ when $g_{i,t}$ and $a_t$ are fixed, and $\eta$ is a small value that meets $\eta < 0.03$.
\end{observation}

This indicates that the aggregated features change with very similar underlying dynamics within days. When $g_{i,t}$ and $a_t$ are fixed, for any two possible values $\hat{g}$ and $\tilde{g}$ of $g_i^{t+1}$, we have:
\begin{equation}  \label{eq:cover}
\begin{aligned} 
\frac {\Arrowvert  \hat{g} - \tilde{g} \Arrowvert ^2 }   {\min(\Arrowvert  \hat{g} \Arrowvert^2, \Arrowvert  \tilde{g} \Arrowvert^2)} & \leq  \left(  \frac{2\eta} {1-\eta} \right)^2  \\
                    &  \leq 0.01 \\
\end{aligned}
\end{equation}
According to Assumption~\ref{assumption: eq_pv}, we can deem any possible value of $g_i^{t+1}$ to be the same, which means $p(g_i' | g_i, A = a) = 1$. According to (\ref{eq:transition_3}), finally we get:
\begin{equation}  \label{eq:transition_4}
p(s, a, s^{'}) = 1
\end{equation}
This means the state transition is consistent among different days, leading to a robust MDP. 

\noindent \textbf{Action Space.}
With established state and transition, we now need to formulate the decision action. Most previous works used reinforcement learning to control the bid directly, so the action is to set bid prices (costs). However, applying the idea to our model would result in setting a batch cost for all the auctions in the period. It is hard to derive impression level bid price and more importantly, this cannot achieve real-time bidding.


Instead of generating bid prices, we take a control-by-model scheme: we deploy a linear approximator as the real-time bidding model to fit the optimal bid prices, and we utilize reinforcement learning to learn an optimal policy to control the real-time bidding model. Hence, the action here is the parameter control of the linear approximator function, rather than the bid decision itself.

Previous studies have shown that the optimal bid price has a linear relationship with the impression-level evaluation (e.g. CTR) \cite{perlich2012bid,lee2012estimating}. In this paper, we adopt the predicted conversion rate (PCVR) as the independent variable in the linear approximator function for real-time bidding which is defined as:
\begin{equation}  \label{eq:rtb}
opt\_bidprice =  f(PCVR) = \alpha \cdot PCVR
\end{equation}

To sum up, the robust MDP we propose is modeled as follows:

\begin{center}
\begin{tabular}{|c|c|}
\hline
state & $<b, t, \vec g>$ \\
\hline
action & set $\alpha$ \\
\hline
reward & $PUR\_AMT$ gained in one step \\
\hline
episode & a single day \\
\hline
\end{tabular}
\end{center}

\label{subsection:prof}

\subsection{Algorithm}
 
In this work, we take a value-based approach as our solution. The goal is to find an optimal policy $\pi(s_t)$ which can be mathematically written as:
\begin{equation} \label{eq:bellman1}
\pi(s_t) =  \mathop{\arg\max}_{a} Q(s_t, a_t)
\end{equation}
Where,
\begin{equation} \label{eq:q_f}
Q(s_t, a_t) =  \mathbb{E} [R_t | S = s_t, A = a_t]
\end{equation}
\begin{equation} \label{eq:rt}
R_t = \sum_{k=0}^{m-t} \gamma^{k} r(s_{t+k}, a_{t+k})
\end{equation}
$R_t$ in Eq.~(\ref{eq:rt}) is the accumulated long term reward that need to be maximized. $Q(s_t, a_t)$ in Eq. (\ref{eq:q_f}) is the standard action value function \cite{sutton1998reinforcement,poole2010artificial} which captures the expected value of $R_t$ given $s_t$ and $a_t$. By finding the optimal $Q$ function for each state $s_t$ iteratively, the agent could derive an optimal sequential decision.

By Bellman equation \cite{sutton1998reinforcement,poole2010artificial}, we could get:
\begin{equation} \label{eq:bellman}
Q^{*}(s_t, a_t) =   \mathbb{E} [ r(s,a)  +  \gamma  \max_{a_{t+1}} Q^{*}(s_{t+1}, a_{t+1}) \arrowvert S = s_t, A = a_t  ]
\end{equation}
Eq. (\ref{eq:bellman}) reveals the relation of $Q^{*}$ values between step $t$ and step $t+1$, where the $Q^*$ value denotes the optimal value for $Q(s,a)$ given $s$ and $a$. For small problems, the optimal action value function can be exactly solved by applying Eq. (\ref{eq:bellman}) iteratively. However, due to the exponential complexity of our model, we adopt a DQN algorithm similar to \cite{mnih2013playing,mnih2015human} which employs a deep neural network (DNN) with weights $\theta$ to approximate $Q$. Besides, we also map the action space into 100 discrete values for decision making. Thus, the deep neural network can be trained by minimizing the loss functions in the following iteratively:
\begin{equation} \label{eq:loss_1}
L_i(\theta_i) =  \mathbb{E} [ (y_i - Q_{train} (s_t, a_t; \theta_i))^{2} ]
\end{equation}

Where, $y_i = r(s,a)  +  \gamma  \max_{a_{t+1}} Q_{target}(s_{t+1}, a_{t+1}) $

The core idea of minimizing the loss in Eq. (\ref{eq:loss_1}) is to find a DNN that can closely approximate the real optimal $Q$ function, using Eq. (\ref{eq:bellman}) as an iterative solver. Similar to \cite{mnih2013playing}, the target network is utilized for stable convergence, which is updated by train network every C steps.

\makeatletter  
\def\BState{\State\hskip-\ALG@thistlm}  
\makeatother  

\begin{algorithm}[h]
  \caption{DQN Learning}
  \label{algorithm:dqn} 
  \begin{algorithmic}[1]  
    \For{$epsode=1$ to $n$}  
      \State Initialize replay memory $D$ to capacity $N$
      \State Initialize action value functions $(Q_{train}, Q_{ep}, Q_{target})$ with weights $\theta_{train}, \theta_{ep}, \theta_{target}$
      \For{$t=1$ to $m$}
         \State With probability $\epsilon$ select a random action $a_t$
         \State otherwise select $a_t = \mathop{\arg\max}_{a} Q_{ep}(s_t, a; \theta_{ep}) $
         \State Execute action $a_t$ to auction simulator and observe state $s_{t+1}$ and reward $r_t$
         \State if budget of $s_{t+1}$ < 0, then continue
         \State Store transition $(s_t, a_t, r_t, s_{t+1})$ in $D$
         \State Sample random mini batch of transitions $(s_j, a_j, r_j, s_{j+1})$ from $D$
         \If{$j = m$}  
               \State Set $y_j = r_j$ 
         \Else  
               \State Set $y_j = r_j + \gamma  \mathop{\arg\max}_{a_{'}} Q_{target}(s_{j+1}, a_{,}, \theta_{target})$ 
         \EndIf
         \State Perform a gradient descent step on the loss function $(y_j - Q_{train}(s_j, a_j; \theta_{train}))^{2}$
      \EndFor
      \State Update $\theta_{ep}$ with $\theta$
      \State Every C steps, update $\theta_{target}$ with $\theta$
    \EndFor  
  \end{algorithmic}  
\end{algorithm}

The corresponding details of the algorithm is presented in Algorithm \ref{algorithm:dqn}. In addition to train network and target network, we also introduce an episode network for better convergence. Algorithm \ref{algorithm:dqn} works in 3 modes to find the optimal strategy: 1) Listening. The agent will record each <$s_t, a_t, r(s_t, a_t), s_{t+1}$> into a replay memory buffer; 2) Training. The agent grabs a mini-batch from the replay memory and performs gradient descent for minimizing the loss in Eq. (\ref{eq:loss_1}). 3) Prediction. The agent will generate an action for the next step greedily by the Q-network. By iteratively performing these 3 modes, an optimal policy could be found.

\label{subsec:algorithm}

\subsection{The Massive-agent Model}

The model in Section \ref{subsec:algorithm} works well when there are only a few agents. However, in the scenario of thousands or millions of agents, the global performance would decrease due to competition. Hence, we proposed an approach for handling the massive-agent problem. 

\begin{figure}
\includegraphics[height=1.2in, width=2.6in]{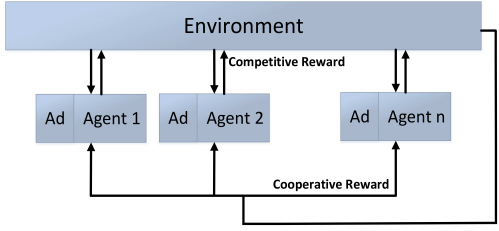}
\caption{Massive-agent Framework}
\label{fig:massive}
\end{figure}

The core idea is to combine the private competitive objective with a public cooperative objective. We designed a cooperative framework: for each ad, we deploy an independent agent to learn the optimal policy according to its own states. The learning algorithm for each agent is very similar to Algorithm~\ref{algorithm:dqn}. The difference is, after all agents made decisions, each agent will receive a competitive feedback representing its own reward and a cooperative feedback representing the global reward of all the agents. The learning framework is illustrated in Figure~\ref{fig:massive}.

\subsection{System Architecture}

\begin{figure}
\includegraphics[height=2in, width=3.2in]{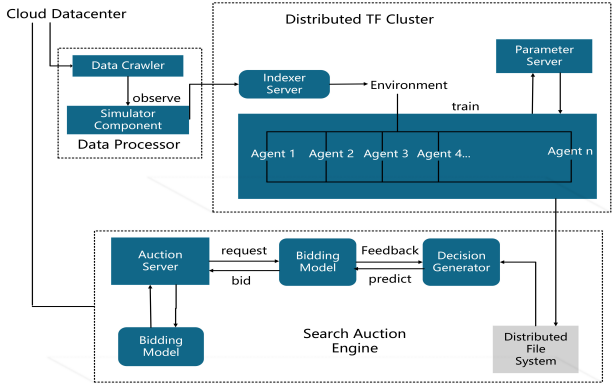}
\caption{System Achitecture}
\label{fig:archi}
\end{figure}

In this subsection, we provide an introduction to the architecture of our system depicted by Figure~\ref{fig:archi}. There are mainly three parts: Data Processor, Distributed Tensor-Flow Cluster, Search Auction Engine. 

\noindent \textbf{Data Processor}
The core component of this part is the simulator, which is in charge of RL exploration. In real search auction platforms, it is usually difficult to obtain precise exploration data. On one hand, we cannot afford to perform many random bidding predictions in online system; on the other hand, it is also hard to generate precise model-based data by pure prediction. For this reason, we build a simulator component for trial and error, which utilizes both model-free data such as real auction logs and model-based data such as predicted conversion rate to generate simulated statistical features for learning. The advantage of the simulator is, auctions with different bid decision could be simulated rather than predicted, since we have the complete auction records for all ads. In particular, we don't need to predict the market price, which is quite hard to predict in SS auction. Besides, for corresponding effected user behavior which can't be simply simulated (purchase, etc), we use a mixed method with simulation and PCVR prediction, With this method, the simulator can generate various effects, e.g. ranking, clicks, costs, etc.

\noindent \textbf{Distributed Tensor-Flow Cluster}
This is a distributed cluster deployed on tensor-flow. The DRL model will be trained here in a distributed manner with parameter servers to coordinate weights of networks. Since DRL usually needs huge amounts of samples and episodes for exploration, and in our scenario thousands of agents need to be trained parallelly, we deployed our model on 1000 CPUs and 40 GPUs, with capability of processing 200 billion sample instances within 2 hours.

\noindent \textbf{Search Auction Engine}
The auction engine is the master component. It sends requests and impression-level features to the bidding model and get bid prices back in real-time. The bidding model, in turn, periodically sends statistical online features to and get from the decision generator optimal policies which are outputted by the trained Q-network.

\label{subsection:sys}

\section{Experimental Setup}
Our methods are tested both by offline evaluation (Section \ref{section:offline}) and via online evaluation (Section \ref{section:online}) on a large e-commerce search auction platform of Alibaba Corporation with real advertisers and auctions. In this section, we introduce the dataset, compared methods and parameter setting.

\subsection{Dataset}
We randomly select 1000 big ads in Alibaba's search auction platform, which on average cover 100 million auctions per day on the platform, for offline/online evaluation. The offline benchmark dataset is extracted from the search auction log for two days of late December, 2017. Each auction instance contains (but not limited to) the bids, the clicks, the auction ranking list and the corresponding predicted features such as PCVR that stand for the predicted utility for a specific impression. For evaluation, we use one day collection as the training data and the other day collection for test. Note that we cannot use the test collection directly for test since the bidding actions have already been made therein. Hence, we perform evaluation by simulation base on the test collection. Both collections contain over 100 million auctions. For online evaluation, a standard A/B test is conducted online. Over 100 million auction instances of the 1000 ads are collected one day in advance for training, and the trained models are used to generate real bidding decisions online.


In order to better evaluate the effectiveness of our single-agent model, we also conduct separate experiments on 10 selected ads with disjoint keyword sets, for both offline and online evaluation. Besides, we use the data processor in our system to generate trial-and-error training data for RL methods. In particular, 200 billion simulated auctions are generated by the simulator (described in Section \ref{subsection:sys}) for training in offline/online evaluation. The simulated datasets are necessary for boosting the performance of RL methods. 


\subsection{Compared Methods and Evaluation Metric}
The compared bidding methods in our experiments include:

\vspace{2mm} \noindent \textbf{Keyword-level bidding (KB)}: KB bids based on keyword-level. It is a simple and efficient online approach adopted by search auction platforms such as the Alibaba's search auction platform. We treat this algorithm as the fundamental baseline of the experiments.

\vspace{2mm} \noindent \textbf{RL with auction-level MDP (AMDP)}: AMDP optimizes sequence bidding decisions by auction-level DRL algorithm \cite{wang2017ladder}. As in \cite{wang2017ladder}, this algorithm samples an auction in every 100 auctions interval as the next state.

\vspace{2mm} \noindent \textbf{RL with robust MDP (RMDP)}: This is the algorithm we proposed in this paper. RMDP is single-agent oriented, without considering the competition between ads.


\vspace{2mm} \noindent \textbf{Massive-agent RL with robust MDP (M-RMDP)}: This is the algorithm extended from RMDP for handling the massive-agent problem.

To evaluate the performance of the algorithms, we use the evaluation metric $PUR\_AMT / COST$ under the same $COST$ constraint. It should be noted that, due to Alibaba's business policy, we temporarily cannot expose the absolute performance values for the KB algorithm. Hence, we use relative improvement values with respect to KB instead. This do not affect performance comparison. We are applying for an official data exposure agreement currently.

\subsection{Hyper-parameter Setting}
To facilitate the implementation of our method, we provide the settings of some key hyper-parameters in Table \ref{tab:hyper}. It is worth to note that, for all agents we use the same hyper-parameter setting and DNN network structure.


\begin{center}
\begin{table} \footnotesize
  \caption{Hyper-parameter settings}
  \label{tab:hyper}
  \begin{tabular}{cc}
    \toprule
    Hyper-parameter & Setting \\
    \midrule
    Target network update period & 10000 steps \\
    Memory size & 1 million \\
    Learning rate & 0.0001 with RMSProp method \cite{tieleman2012lecture}  \\
    Batch size & 300 \\
    Network structure & 4-layer DNN \\
    DNN layer sizes & [15, 300, 200, 100] \\
  \bottomrule
\end{tabular}
\end{table}
\end{center}

\section{Offline Evaluation}
The purpose of this experiment is to answer the following questions. (i) How does the DRL algorithm works for search auction data? (ii) Does RMDP outperform AMDP under changing environment? (iii) IS the multi-agent problem well handled by M-RMDP?

\label{section:offline}

\subsection{Single-agent Analysis}

The performance comparison in terms of $PUR\_AMT / COST$ is presented in Table~\ref{tab:single_agent_offline}, where all the algorithms are compared under the same cost constraint. Thus, the performance in Table~\ref{tab:single_agent_offline} actually depicts the capability of the bidding algorithm in obtaining more $PUR\_AMT$ under same cost, compared to KB. It shows that, on the test data RMDP outperforms KB and AMDP. However, if we compare the performance on the training dataset, the AMDP algorithm is the best since it models decision control at the auction level. Nevertheless, AMDP performs poorly on the test data, indicating serious overfitting to the training data. This result demonstrates that the auction sequences cannot ensure a consistent transition probability distribution between different days. In contrast, RMDP shows stable performance between training and test, which suggests that RMDP indeed captures consistent transition patterns under environment changing by hour-aggregation of auctions. 

Besides, Table~\ref{tab:single_agent_offline} also shows that for each of the 10 ads RMDP consistently learns a much better policy that KB. Since we use the same hyper-parameter setting and network structure, it indicates the performance of our method is not very sensitive to hyper-parameter settings.

Furthermore, the results showed in Table \ref{tab:single_agent_offline} illuminate us a general thought about reinforcement learning in the online advertising field. The power of reinforcement learning is due to sequential decision making. Normally, the more frequently the model can get feedback and adjust its control policy, the better the performance will be (AMDP on the training data). However, in the scenario of progressively changing environment, a frequent feedback information might contain too much stochastic noise. A promising solution for robust training is to collect statistical data with proper aggregation. Nevertheless, the drawback of this technique is sacrificing adjust frequency. Generally, a good approach of DRL for online advertising is actually the consequence of a good trade-off between feedback frequency and data trustworthiness (higher aggregation levels exhibit more consistent transition patterns and thus are more trustworthy).


\begin{table} \footnotesize
  \caption{Results for the offline single-agent case. Performance is relative improvement of $PUR\_AMT / COST$ compared to KB.}
  \label{tab:single_agent_offline}
  \begin{tabular}{p{1cm}p{1.1cm}p{1.1cm}p{1.3cm}p{1.3cm}}
    \toprule
    ad\_id & AMDP(train) & AMDP(test) & RMDP(train) & RMDP(test) \\
    \midrule
    740053750 & 334.28\% & 5.56\% & 158.81\% & 136.86\% \\
    75694893 & 297.27\% & -4.62\% & 95.80\% & 62.18\%  \\
    749798178 & 68.89\% & 8.04\% & 38.54\% & 34.14\% \\
    781346990 & 227.91\% & -20.08\% & 79.52\% & 57.99\% \\
    781625444 & 144.93\% & -72.46\% & 53.62\% & 38.79\%  \\
    783136763 & 489.09\% & 38.18\% & 327.88\% & 295.76\% \\
    750569395 & 195.42\% & -15.09\% & 130.46\% & 114.29\% \\
    787215770 & 253.64\% & -41.06\% & 175.50\% & 145.70\% \\
    802779226 & 158.50\% & -44.67\% & 79.07\% & 72.71\% \\
    805113454 & 510.13\% &  -8.86\% & 236.08\% & 195.25\% \\
    Avg. & 250.03\% & -10.06\% & 120.01\% & 98.74\% \\
  \bottomrule
\end{tabular}
\end{table}

\subsection{Multi-agent Analysis}
To investigate how DRL works with many agents competing with each other, we run KB, RMDP and M-RMDP on all the 1000 ads of the offline dataset. In this experiment and the following online experiments, we do not involve AMDP since it has already been shown to perform even worse than KB in the single-agent case.

\begin{center}
\begin{figure}
\subfigure[$COST/COST_{KB}$]
{ 
\begin{minipage}[t]{0.4\linewidth}
\includegraphics[height=30mm, width=40mm]{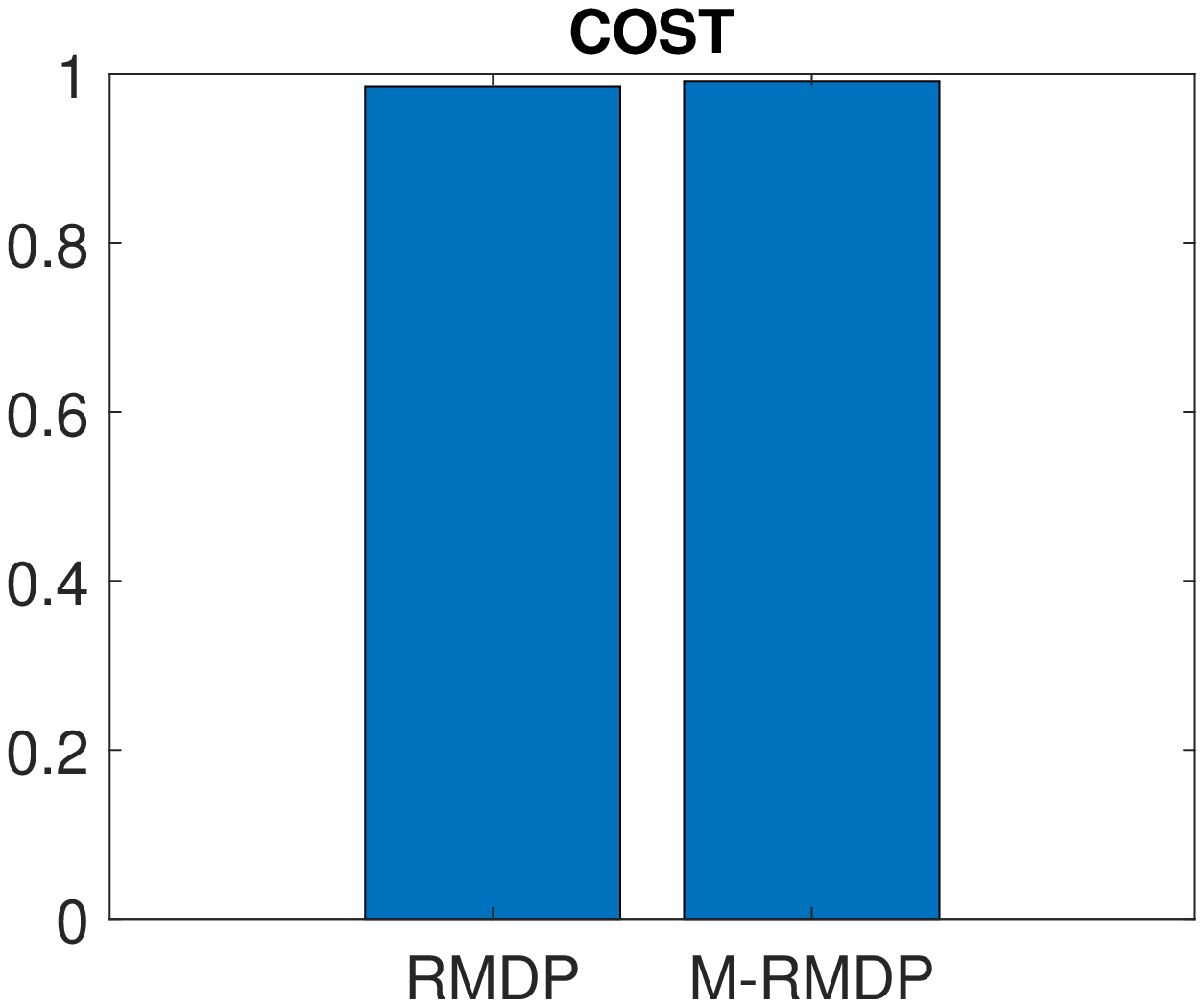}
\end{minipage}
}
\subfigure[$PUR\_AMT/COST$]
{ 
\begin{minipage}[t]{0.4\linewidth}
\includegraphics[height=30mm, width=40mm]{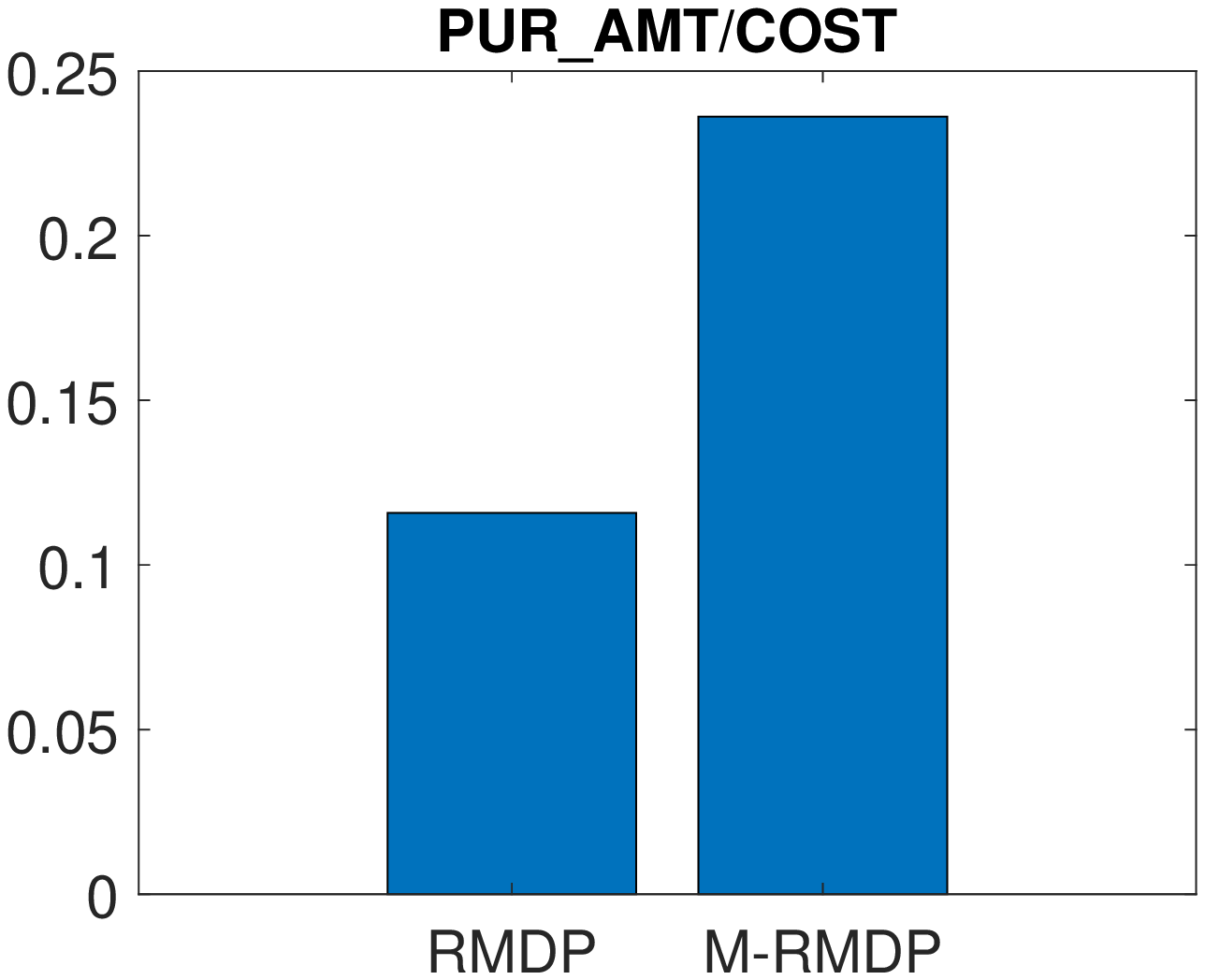}
\end{minipage}
}
\caption{Performance on 1000 ads (offline).}
\label{fig:expoffline}
\end{figure}
\end{center}

The averaged results for the 1000 ads are shown in Figure~\ref{fig:expoffline}. The first thing we can observe is that, the costs of all the algorithms are similar (note that the y-axis of Figure~\ref{fig:expoffline}(a) measures the ratio to KB's cost), while the $PUR\_AMT / COST$ results are different. It shows that RMDP still outperforms KB, but the relative improvement is not as high as in the single-agent case. Compared to RMDP, M-RMDP shows a prominent improvement in terms of $PUR\_AMT / COST$. This consolidates our speculation that by equipping each agent with a cooperative objective, the performance could be enhanced.

\section{Online evaluation}
This section presents the results of online evaluation in the real-world auction environment of the Alibaba search auction platform with a standard A/B testing configuration. As all the results are collected online, in addition to the key performance metric $PUR\_AMT$ / $COST$, we also show results on metrics including conversion rate (CVR), return on investment (ROI) and cost per click (PPC). These metrics, though different, are related to our objective.

\label{section:online}

\subsection{Single-agent Analysis}
We continue to use the same 10 ads for evaluation. The detailed performance of RMDP for each ad is listed in Table~\ref{tab:single_agent_online}. It can be observed from the last line that the performance of RMDP outperforms that of KB with an average of 35.04\% improvement of $PUR\_AMT / COST$. This is expected as the results are similar to those of the offline experiment. Although the percentages of improvement are different between online and offline cases, this is also intuitive since we use simulated results based on the test collection for evaluation in the offline case. This suggests our RMDP model is indeed robust when deployed to real-world auction platforms. Besides, we also take other related metrics as a reference. We find that there is an average of 23.7\% improvement in CVR, an average of 21.38\% improvement in ROI and a slight average improvement (5.16\%) in PPC (the lower, the better). This means our model could also indirectly improve other correlated performance metrics that are usually considered in online advertising. The slight improvement in PPC means that we help advertisers save a little of his/her cost per click, although not prominent.

\begin{table} \footnotesize
  \caption{Results for the online single-agent case. Performance is relative improvement of RMDP compared to KB.}
  \label{tab:single_agent_online}
  \begin{tabular}{ccccl}
    \toprule
    ad\_id & $PUR\_AMT / COST$ & CVR & ROI & PPC \\
    \midrule
    740053750 & 65.19\% & 60.78\% & 19.01\% & -2.67\%  \\
    75694893  & 23.59\% & 8.83\%  & 12.75\% &  -11.94\% \\
    749798178 & 4.15\% & -7.98\%  & -0.63\% &  -11.66\%\\
    781346990 & 41.6\% &  49.12\%  & 43.86 \% &  5.33\% \\
    781625444 & -9.79\% &  30.95\% & 7.73\%  &  14.28\%\\
    783136763 & 55.853\% &  27.03\% & 52.87\% &  -18.49\% \\
    750569395 & 2.854\% & 1.65\%  & 19.60\%  &   8.81\%\\
    787215770 & 21.52\% &  32.97\% & 46.94\% &   -8.61\% \\
    802779226 & 31.44\% & 46.93\%  & 19.97\%  & 11.78\%\\
    805113454 & 57.08\% &  78.64\%  & 68.73\%  &  13.74\%\\
    Avg. &   35.04\% & 23.11\% &  21.38\% & -5.16\%  \\
  \bottomrule
\end{tabular}
\end{table}

\subsection{Multi-agent Analysis}
A standard online A/B test on the 1000 ads is carried out on Feb. 5th, 2018. The averaged relative improvement results of RMDP and M-RMDP compared to KB are depicted in Table \ref{tab:massive_agent}. We can see that, similar to the offline experiment, M-RMDP outperforms the online KB algorithm and RMDP in several aspects: i) higher $PUR\_AMT / COST$, which is the objective of our model; ii) higher ROI and CVR, which are related key utilities that advertisers concern. It is again observed that PPC is slightly improved, which means that our model can slightly help advertisers save their cost per click. The performance improvement of RMDP is lower that that in the online single-agent case (Table~\ref{tab:single_agent_online}). The reason could be that the competition among the ads affect its performance. In comparison, M-RMDP can well handle the multi-agent problem.

\begin{table} \footnotesize
  \caption{Results for the online multi-agent case. Performance is relative improvement compared to KB.}
  \label{tab:massive_agent}
  \begin{tabular}{cccccl}
    \toprule
    Algorithm & $PUR\_AMT / COST$ & ROI & CVR & PPC \\
    \midrule
    RMDP & 6.29\% & 26.51\% & 3.12\% & -3.36\% \\
    M-RMDP & 13.01\%  &  39.12\% & 12.62\% & -0.74\% \\
  \bottomrule
\end{tabular}
\end{table}

\subsection{Convergence Analysis} 
\begin{center}
\begin{figure}
\subfigure[]
{ 
\begin{minipage}[t]{0.45\linewidth}
\includegraphics[height=30mm, width=40mm]{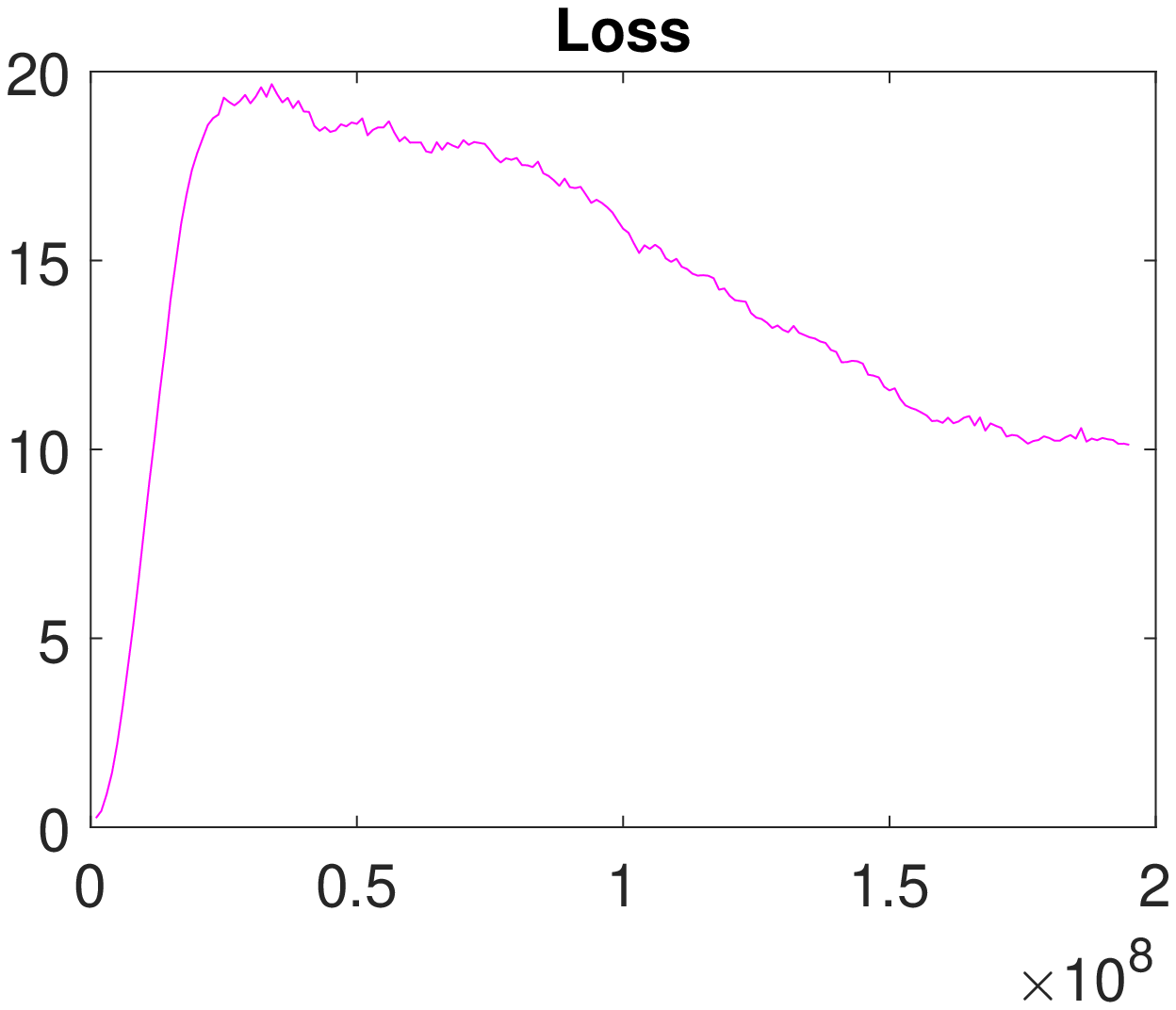}
\end{minipage}
}
\subfigure[]
{ 
\begin{minipage}[t]{0.45\linewidth}
\includegraphics[height=30mm, width=40mm]{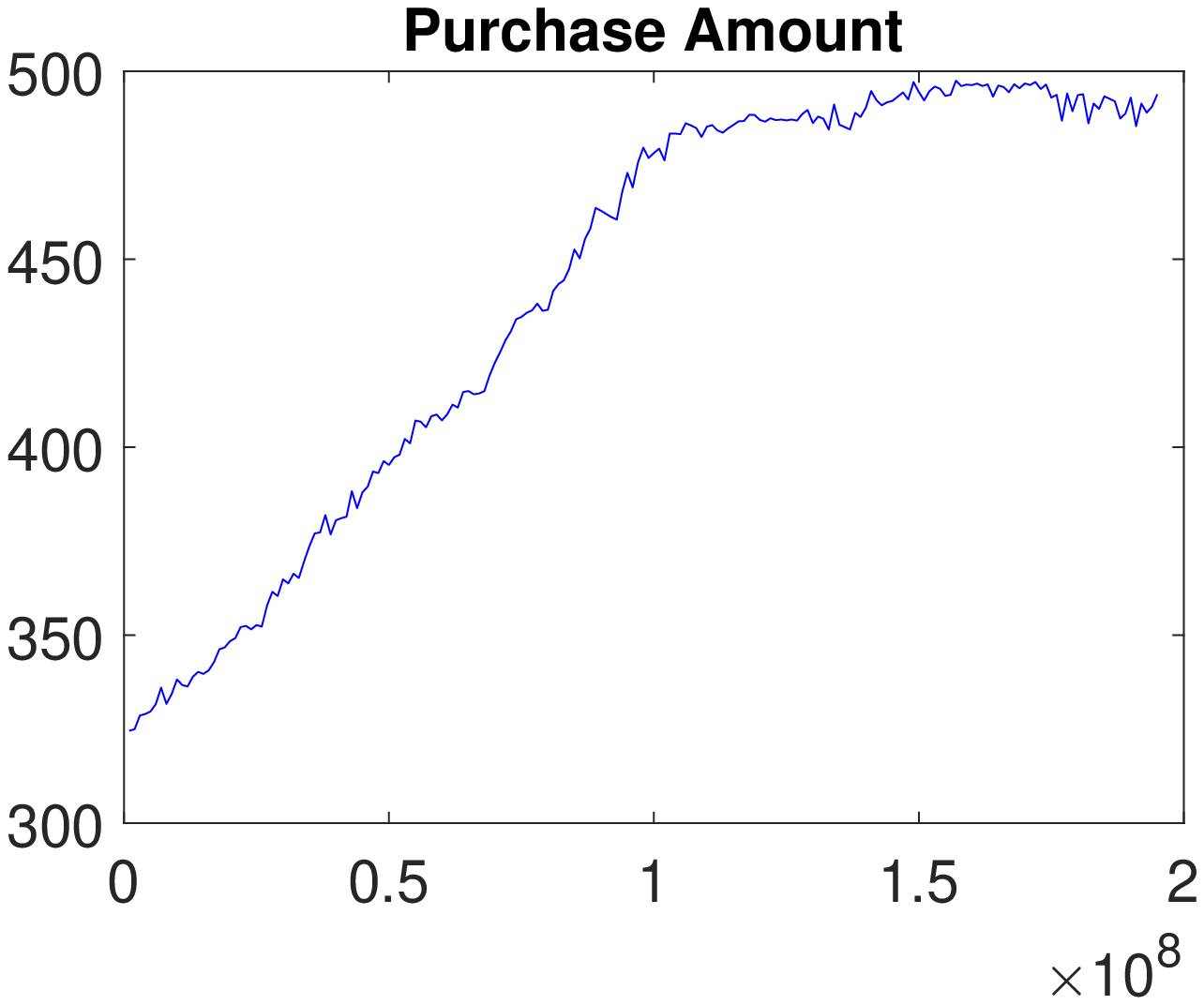}
\end{minipage}
}
\subfigure[]
{ 
\begin{minipage}[t]{0.45\linewidth}
\includegraphics[height=30mm, width=40mm]{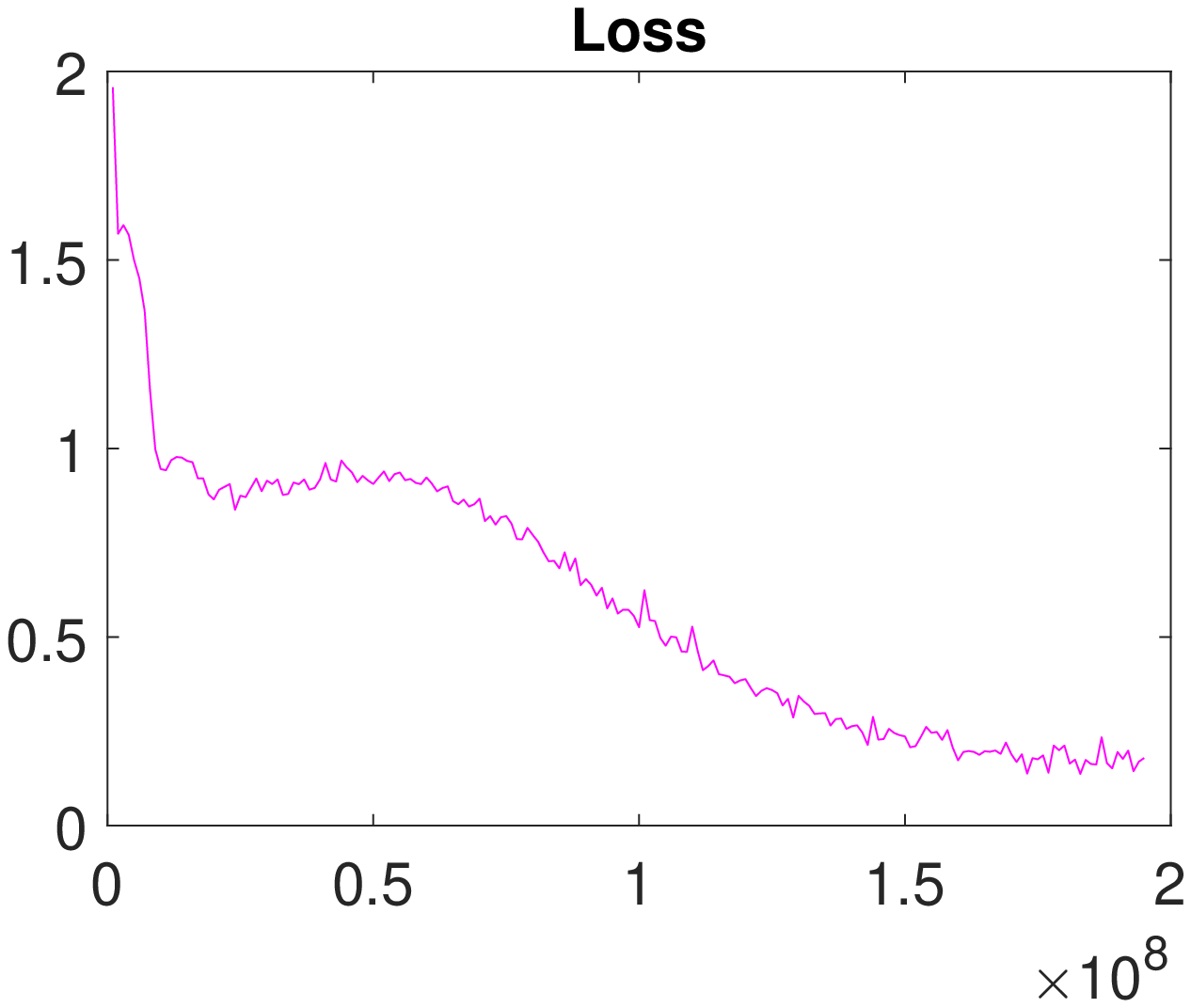}
\end{minipage}
}
\subfigure[]
{ 
\begin{minipage}[t]{0.45\linewidth}
\includegraphics[height=30mm, width=40mm]{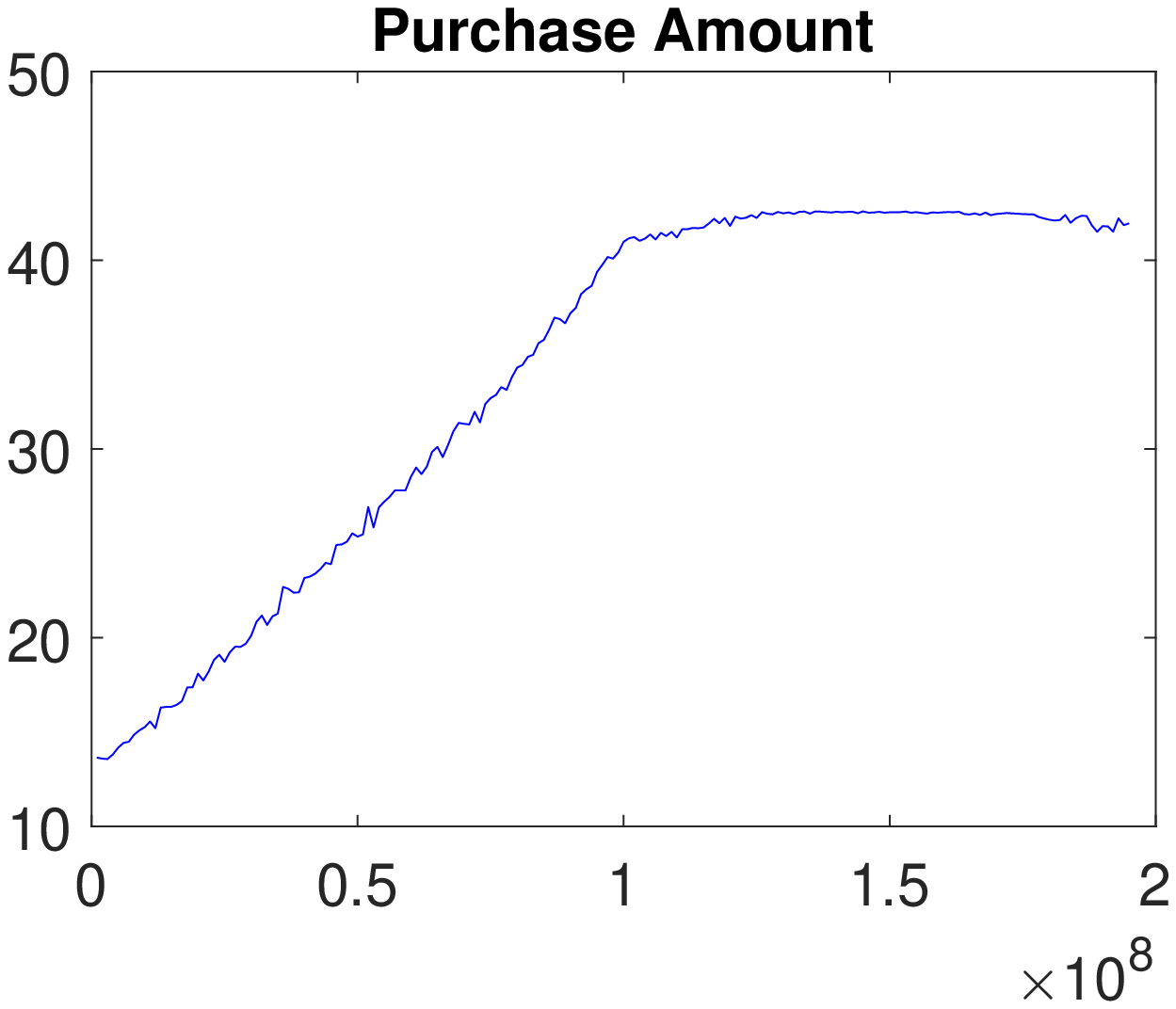}
\end{minipage}
}
\caption{convergence Performance.}
\label{fig:converge}
\end{figure}
\end{center}

We provide convergence analysis of the RMDP model by two example ads in Figure~\ref{fig:converge}. Figures~\ref{fig:converge}(a) and~\ref{fig:converge}(c) show the $Loss$ (i.e. Eq.~(\ref{eq:loss_1})) curves with the number of learning batches processed. Figures~\ref{fig:converge}(b) and~\ref{fig:converge}(d) present the $PUR\_AMT$ (i.e. our optimization objective in Eq.~(\ref{eq:obj_2})) curves accordingly. We observe that, in Figures~\ref{fig:converge}(a) and~\ref{fig:converge}(c) the $Loss$ starts as or quickly increases to a large value and then slowly converge to a much smaller value, while in the same batch range $PUR\_AMT$ improves persistently and becomes stable (Figures~\ref{fig:converge}(b) and~\ref{fig:converge}(d)). This provides us a good evidence that our DQN algorithm has a solid capability to adjust from a random policy to an optimal solution. In our system, the random probability $\epsilon$ of exploration in Algorithm~\ref{algorithm:dqn} was initially set to $1.0$, and decays to 0.0 during the learning process. The curves in Figure \ref{fig:converge} demonstrate a good convergence performance of RL.

Besides, we observe that the loss value converges to a relatively small value after about 150 million sample batches are processed, which suggests that the DQN algorithm is data expensive. It needs large amounts of data and episodes to find an optimal policy.

\bibliographystyle{ACM-Reference-Format}
\bibliography{sample-bibliography}


\begin{thebibliography}{00}


\ifx \showCODEN    \undefined \def \showCODEN     #1{\unskip}     \fi
\ifx \showDOI      \undefined \def \showDOI       #1{#1}\fi
\ifx \showISBNx    \undefined \def \showISBNx     #1{\unskip}     \fi
\ifx \showISBNxiii \undefined \def \showISBNxiii  #1{\unskip}     \fi
\ifx \showISSN     \undefined \def \showISSN      #1{\unskip}     \fi
\ifx \showLCCN     \undefined \def \showLCCN      #1{\unskip}     \fi
\ifx \shownote     \undefined \def \shownote      #1{#1}          \fi
\ifx \showarticletitle \undefined \def \showarticletitle #1{#1}   \fi
\ifx \showURL      \undefined \def \showURL       {\relax}        \fi
\providecommand\bibfield[2]{#2}
\providecommand\bibinfo[2]{#2}
\providecommand\natexlab[1]{#1}
\providecommand\showeprint[2][]{arXiv:#2}

\bibitem[\protect\citeauthoryear{Amin, Kearns, Key, and Schwaighofer}{Amin
  et~al\mbox{.}}{2012}]%
        {amin2012budget}
\bibfield{author}{\bibinfo{person}{Kareem Amin}, \bibinfo{person}{Michael
  Kearns}, \bibinfo{person}{Peter Key}, {and} \bibinfo{person}{Anton
  Schwaighofer}.} \bibinfo{year}{2012}\natexlab{}.
\newblock \showarticletitle{Budget optimization for sponsored search: Censored
  learning in MDPs}.
\newblock \bibinfo{journal}{{\em arXiv preprint arXiv:1210.4847\/}}
  (\bibinfo{year}{2012}).
\newblock


\bibitem[\protect\citeauthoryear{Borgs, Chayes, Immorlica, Jain, Etesami, and
  Mahdian}{Borgs et~al\mbox{.}}{2007}]%
        {borgs2007dynamics}
\bibfield{author}{\bibinfo{person}{Christian Borgs}, \bibinfo{person}{Jennifer
  Chayes}, \bibinfo{person}{Nicole Immorlica}, \bibinfo{person}{Kamal Jain},
  \bibinfo{person}{Omid Etesami}, {and} \bibinfo{person}{Mohammad Mahdian}.}
  \bibinfo{year}{2007}\natexlab{}.
\newblock \showarticletitle{Dynamics of bid optimization in online
  advertisement auctions}. In \bibinfo{booktitle}{{\em Proceedings of the 16th
  international conference on World Wide Web}}. ACM, \bibinfo{pages}{531--540}.
\newblock


\bibitem[\protect\citeauthoryear{Borgs, Chayes, Immorlica, Mahdian, and
  Saberi}{Borgs et~al\mbox{.}}{2005}]%
        {borgs2005multi}
\bibfield{author}{\bibinfo{person}{Christian Borgs}, \bibinfo{person}{Jennifer
  Chayes}, \bibinfo{person}{Nicole Immorlica}, \bibinfo{person}{Mohammad
  Mahdian}, {and} \bibinfo{person}{Amin Saberi}.}
  \bibinfo{year}{2005}\natexlab{}.
\newblock \showarticletitle{Multi-unit auctions with budget-constrained
  bidders}. In \bibinfo{booktitle}{{\em Proceedings of the 6th ACM conference
  on Electronic commerce}}. ACM, \bibinfo{pages}{44--51}.
\newblock


\bibitem[\protect\citeauthoryear{Broder, Gabrilovich, Josifovski, Mavromatis,
  and Smola}{Broder et~al\mbox{.}}{2011}]%
        {broder2011bid}
\bibfield{author}{\bibinfo{person}{Andrei Broder}, \bibinfo{person}{Evgeniy
  Gabrilovich}, \bibinfo{person}{Vanja Josifovski}, \bibinfo{person}{George
  Mavromatis}, {and} \bibinfo{person}{Alex Smola}.}
  \bibinfo{year}{2011}\natexlab{}.
\newblock \showarticletitle{Bid generation for advanced match in sponsored
  search}. In \bibinfo{booktitle}{{\em Proceedings of the fourth ACM
  international conference on Web search and data mining}}. ACM,
  \bibinfo{pages}{515--524}.
\newblock


\bibitem[\protect\citeauthoryear{Busoniu, Babuska, and De~Schutter}{Busoniu
  et~al\mbox{.}}{2008}]%
        {busoniu2008comprehensive}
\bibfield{author}{\bibinfo{person}{Lucian Busoniu}, \bibinfo{person}{Robert
  Babuska}, {and} \bibinfo{person}{Bart De~Schutter}.}
  \bibinfo{year}{2008}\natexlab{}.
\newblock \showarticletitle{A comprehensive survey of multiagent reinforcement
  learning}.
\newblock \bibinfo{journal}{{\em IEEE Trans. Systems, Man, and Cybernetics,
  Part C\/}} \bibinfo{volume}{38}, \bibinfo{number}{2} (\bibinfo{year}{2008}),
  \bibinfo{pages}{156--172}.
\newblock


\bibitem[\protect\citeauthoryear{Cai, Ren, Zhang, Malialis, Wang, Yu, and
  Guo}{Cai et~al\mbox{.}}{2017}]%
        {cai2017real}
\bibfield{author}{\bibinfo{person}{Han Cai}, \bibinfo{person}{Kan Ren},
  \bibinfo{person}{Weinan Zhang}, \bibinfo{person}{Kleanthis Malialis},
  \bibinfo{person}{Jun Wang}, \bibinfo{person}{Yong Yu}, {and}
  \bibinfo{person}{Defeng Guo}.} \bibinfo{year}{2017}\natexlab{}.
\newblock \showarticletitle{Real-time bidding by reinforcement learning in
  display advertising}. In \bibinfo{booktitle}{{\em Proceedings of the Tenth
  ACM International Conference on Web Search and Data Mining}}. ACM,
  \bibinfo{pages}{661--670}.
\newblock


\bibitem[\protect\citeauthoryear{Chen, Berkhin, Anderson, and Devanur}{Chen
  et~al\mbox{.}}{2011}]%
        {chen2011real}
\bibfield{author}{\bibinfo{person}{Ye Chen}, \bibinfo{person}{Pavel Berkhin},
  \bibinfo{person}{Bo Anderson}, {and} \bibinfo{person}{Nikhil~R Devanur}.}
  \bibinfo{year}{2011}\natexlab{}.
\newblock \showarticletitle{Real-time bidding algorithms for performance-based
  display ad allocation}. In \bibinfo{booktitle}{{\em Proceedings of the 17th
  ACM SIGKDD international conference on Knowledge discovery and data mining}}.
  ACM, \bibinfo{pages}{1307--1315}.
\newblock


\bibitem[\protect\citeauthoryear{Even~Dar, Mirrokni, Muthukrishnan, Mansour,
  and Nadav}{Even~Dar et~al\mbox{.}}{2009}]%
        {even2009bid}
\bibfield{author}{\bibinfo{person}{Eyal Even~Dar}, \bibinfo{person}{Vahab~S
  Mirrokni}, \bibinfo{person}{S Muthukrishnan}, \bibinfo{person}{Yishay
  Mansour}, {and} \bibinfo{person}{Uri Nadav}.}
  \bibinfo{year}{2009}\natexlab{}.
\newblock \showarticletitle{Bid optimization for broad match ad auctions}. In
  \bibinfo{booktitle}{{\em Proceedings of the 18th international conference on
  World wide web}}. ACM, \bibinfo{pages}{231--240}.
\newblock


\bibitem[\protect\citeauthoryear{Feldman, Muthukrishnan, Pal, and
  Stein}{Feldman et~al\mbox{.}}{2007}]%
        {feldman2007budget}
\bibfield{author}{\bibinfo{person}{Jon Feldman}, \bibinfo{person}{S
  Muthukrishnan}, \bibinfo{person}{Martin Pal}, {and} \bibinfo{person}{Cliff
  Stein}.} \bibinfo{year}{2007}\natexlab{}.
\newblock \showarticletitle{Budget optimization in search-based advertising
  auctions}. In \bibinfo{booktitle}{{\em Proceedings of the 8th ACM conference
  on Electronic commerce}}. ACM, \bibinfo{pages}{40--49}.
\newblock


\bibitem[\protect\citeauthoryear{Fuxman, Tsaparas, Achan, and Agrawal}{Fuxman
  et~al\mbox{.}}{2008}]%
        {fuxman2008using}
\bibfield{author}{\bibinfo{person}{Ariel Fuxman}, \bibinfo{person}{Panayiotis
  Tsaparas}, \bibinfo{person}{Kannan Achan}, {and} \bibinfo{person}{Rakesh
  Agrawal}.} \bibinfo{year}{2008}\natexlab{}.
\newblock \showarticletitle{Using the wisdom of the crowds for keyword
  generation}. In \bibinfo{booktitle}{{\em Proceedings of the 17th
  international conference on World Wide Web}}. ACM, \bibinfo{pages}{61--70}.
\newblock


\bibitem[\protect\citeauthoryear{Gu, Lillicrap, Sutskever, and Levine}{Gu
  et~al\mbox{.}}{2016}]%
        {gu2016continuous}
\bibfield{author}{\bibinfo{person}{Shixiang Gu}, \bibinfo{person}{Timothy
  Lillicrap}, \bibinfo{person}{Ilya Sutskever}, {and} \bibinfo{person}{Sergey
  Levine}.} \bibinfo{year}{2016}\natexlab{}.
\newblock \showarticletitle{Continuous deep q-learning with model-based
  acceleration}. In \bibinfo{booktitle}{{\em International Conference on
  Machine Learning}}. \bibinfo{pages}{2829--2838}.
\newblock


\bibitem[\protect\citeauthoryear{Hafner and Riedmiller}{Hafner and
  Riedmiller}{2011}]%
        {hafner2011reinforcement}
\bibfield{author}{\bibinfo{person}{Roland Hafner} {and} \bibinfo{person}{Martin
  Riedmiller}.} \bibinfo{year}{2011}\natexlab{}.
\newblock \showarticletitle{Reinforcement learning in feedback control}.
\newblock \bibinfo{journal}{{\em Machine learning\/}} \bibinfo{volume}{84},
  \bibinfo{number}{1-2} (\bibinfo{year}{2011}), \bibinfo{pages}{137--169}.
\newblock


\bibitem[\protect\citeauthoryear{Kitts and Leblanc}{Kitts and Leblanc}{2004}]%
        {kitts2004optimal}
\bibfield{author}{\bibinfo{person}{Brendan Kitts} {and}
  \bibinfo{person}{Benjamin Leblanc}.} \bibinfo{year}{2004}\natexlab{}.
\newblock \showarticletitle{Optimal bidding on keyword auctions}.
\newblock \bibinfo{journal}{{\em Electronic Markets\/}} \bibinfo{volume}{14},
  \bibinfo{number}{3} (\bibinfo{year}{2004}), \bibinfo{pages}{186--201}.
\newblock


\bibitem[\protect\citeauthoryear{Lee, Jalali, and Dasdan}{Lee
  et~al\mbox{.}}{2013}]%
        {lee2013real}
\bibfield{author}{\bibinfo{person}{Kuang-Chih Lee}, \bibinfo{person}{Ali
  Jalali}, {and} \bibinfo{person}{Ali Dasdan}.}
  \bibinfo{year}{2013}\natexlab{}.
\newblock \showarticletitle{Real time bid optimization with smooth budget
  delivery in online advertising}. In \bibinfo{booktitle}{{\em Proceedings of
  the Seventh International Workshop on Data Mining for Online Advertising}}.
  ACM, \bibinfo{pages}{1}.
\newblock


\bibitem[\protect\citeauthoryear{Lee, Orten, Dasdan, and Li}{Lee
  et~al\mbox{.}}{2012}]%
        {lee2012estimating}
\bibfield{author}{\bibinfo{person}{Kuang-chih Lee}, \bibinfo{person}{Burkay
  Orten}, \bibinfo{person}{Ali Dasdan}, {and} \bibinfo{person}{Wentong Li}.}
  \bibinfo{year}{2012}\natexlab{}.
\newblock \showarticletitle{Estimating conversion rate in display advertising
  from past erformance data}. In \bibinfo{booktitle}{{\em Proceedings of the
  18th ACM SIGKDD international conference on Knowledge discovery and data
  mining}}. ACM, \bibinfo{pages}{768--776}.
\newblock


\bibitem[\protect\citeauthoryear{Levine and Abbeel}{Levine and Abbeel}{2014}]%
        {levine2014learning}
\bibfield{author}{\bibinfo{person}{Sergey Levine} {and} \bibinfo{person}{Pieter
  Abbeel}.} \bibinfo{year}{2014}\natexlab{}.
\newblock \showarticletitle{Learning neural network policies with guided policy
  search under unknown dynamics}. In \bibinfo{booktitle}{{\em Advances in
  Neural Information Processing Systems}}. \bibinfo{pages}{1071--1079}.
\newblock


\bibitem[\protect\citeauthoryear{Mnih, Kavukcuoglu, Silver, Graves, Antonoglou,
  Wierstra, and Riedmiller}{Mnih et~al\mbox{.}}{2013}]%
        {mnih2013playing}
\bibfield{author}{\bibinfo{person}{Volodymyr Mnih}, \bibinfo{person}{Koray
  Kavukcuoglu}, \bibinfo{person}{David Silver}, \bibinfo{person}{Alex Graves},
  \bibinfo{person}{Ioannis Antonoglou}, \bibinfo{person}{Daan Wierstra}, {and}
  \bibinfo{person}{Martin Riedmiller}.} \bibinfo{year}{2013}\natexlab{}.
\newblock \showarticletitle{Playing atari with deep reinforcement learning}.
\newblock \bibinfo{journal}{{\em arXiv preprint arXiv:1312.5602\/}}
  (\bibinfo{year}{2013}).
\newblock


\bibitem[\protect\citeauthoryear{Mnih, Kavukcuoglu, Silver, Rusu, Veness,
  Bellemare, Graves, Riedmiller, Fidjeland, Ostrovski, et~al\mbox{.}}{Mnih
  et~al\mbox{.}}{2015}]%
        {mnih2015human}
\bibfield{author}{\bibinfo{person}{Volodymyr Mnih}, \bibinfo{person}{Koray
  Kavukcuoglu}, \bibinfo{person}{David Silver}, \bibinfo{person}{Andrei~A
  Rusu}, \bibinfo{person}{Joel Veness}, \bibinfo{person}{Marc~G Bellemare},
  \bibinfo{person}{Alex Graves}, \bibinfo{person}{Martin Riedmiller},
  \bibinfo{person}{Andreas~K Fidjeland}, \bibinfo{person}{Georg Ostrovski},
  {et~al\mbox{.}}} \bibinfo{year}{2015}\natexlab{}.
\newblock \showarticletitle{Human-level control through deep reinforcement
  learning}.
\newblock \bibinfo{journal}{{\em Nature\/}} \bibinfo{volume}{518},
  \bibinfo{number}{7540} (\bibinfo{year}{2015}), \bibinfo{pages}{529}.
\newblock


\bibitem[\protect\citeauthoryear{Muthukrishnan, P{\'a}l, and
  Svitkina}{Muthukrishnan et~al\mbox{.}}{2007}]%
        {muthukrishnan2007stochastic}
\bibfield{author}{\bibinfo{person}{S Muthukrishnan}, \bibinfo{person}{Martin
  P{\'a}l}, {and} \bibinfo{person}{Zoya Svitkina}.}
  \bibinfo{year}{2007}\natexlab{}.
\newblock \showarticletitle{Stochastic models for budget optimization in
  search-based advertising}. In \bibinfo{booktitle}{{\em International Workshop
  on Web and Internet Economics}}. Springer, \bibinfo{pages}{131--142}.
\newblock


\bibitem[\protect\citeauthoryear{Perlich, Dalessandro, Hook, Stitelman, Raeder,
  and Provost}{Perlich et~al\mbox{.}}{2012}]%
        {perlich2012bid}
\bibfield{author}{\bibinfo{person}{Claudia Perlich}, \bibinfo{person}{Brian
  Dalessandro}, \bibinfo{person}{Rod Hook}, \bibinfo{person}{Ori Stitelman},
  \bibinfo{person}{Troy Raeder}, {and} \bibinfo{person}{Foster Provost}.}
  \bibinfo{year}{2012}\natexlab{}.
\newblock \showarticletitle{Bid optimizing and inventory scoring in targeted
  online advertising}. In \bibinfo{booktitle}{{\em Proceedings of the 18th ACM
  SIGKDD international conference on Knowledge discovery and data mining}}.
  ACM, \bibinfo{pages}{804--812}.
\newblock


\bibitem[\protect\citeauthoryear{Poole and Mackworth}{Poole and
  Mackworth}{2010}]%
        {poole2010artificial}
\bibfield{author}{\bibinfo{person}{David~L Poole} {and} \bibinfo{person}{Alan~K
  Mackworth}.} \bibinfo{year}{2010}\natexlab{}.
\newblock \bibinfo{booktitle}{{\em Artificial Intelligence: foundations of
  computational agents}}.
\newblock \bibinfo{publisher}{Cambridge University Press}.
\newblock


\bibitem[\protect\citeauthoryear{Schwartz}{Schwartz}{2014}]%
        {schwartz2014multi}
\bibfield{author}{\bibinfo{person}{Howard~M Schwartz}.}
  \bibinfo{year}{2014}\natexlab{}.
\newblock \bibinfo{booktitle}{{\em Multi-agent machine learning: A
  reinforcement approach}}.
\newblock \bibinfo{publisher}{John Wiley \& Sons}.
\newblock


\bibitem[\protect\citeauthoryear{Silver, Huang, Maddison, Guez, Sifre, Van
  Den~Driessche, Schrittwieser, Antonoglou, Panneershelvam, Lanctot,
  et~al\mbox{.}}{Silver et~al\mbox{.}}{2016}]%
        {silver2016mastering}
\bibfield{author}{\bibinfo{person}{David Silver}, \bibinfo{person}{Aja Huang},
  \bibinfo{person}{Chris~J Maddison}, \bibinfo{person}{Arthur Guez},
  \bibinfo{person}{Laurent Sifre}, \bibinfo{person}{George Van Den~Driessche},
  \bibinfo{person}{Julian Schrittwieser}, \bibinfo{person}{Ioannis Antonoglou},
  \bibinfo{person}{Veda Panneershelvam}, \bibinfo{person}{Marc Lanctot},
  {et~al\mbox{.}}} \bibinfo{year}{2016}\natexlab{}.
\newblock \showarticletitle{Mastering the game of Go with deep neural networks
  and tree search}.
\newblock \bibinfo{journal}{{\em nature\/}} \bibinfo{volume}{529},
  \bibinfo{number}{7587} (\bibinfo{year}{2016}), \bibinfo{pages}{484--489}.
\newblock


\bibitem[\protect\citeauthoryear{Sutton and Barto}{Sutton and Barto}{1998}]%
        {sutton1998reinforcement}
\bibfield{author}{\bibinfo{person}{Richard~S Sutton} {and}
  \bibinfo{person}{Andrew~G Barto}.} \bibinfo{year}{1998}\natexlab{}.
\newblock \bibinfo{booktitle}{{\em Reinforcement learning: An introduction}}.
  Vol.~\bibinfo{volume}{1}.
\newblock \bibinfo{publisher}{MIT press Cambridge}.
\newblock


\bibitem[\protect\citeauthoryear{Tampuu, Matiisen, Kodelja, Kuzovkin, Korjus,
  Aru, Aru, and Vicente}{Tampuu et~al\mbox{.}}{2017}]%
        {tampuu2017multiagent}
\bibfield{author}{\bibinfo{person}{Ardi Tampuu}, \bibinfo{person}{Tambet
  Matiisen}, \bibinfo{person}{Dorian Kodelja}, \bibinfo{person}{Ilya Kuzovkin},
  \bibinfo{person}{Kristjan Korjus}, \bibinfo{person}{Juhan Aru},
  \bibinfo{person}{Jaan Aru}, {and} \bibinfo{person}{Raul Vicente}.}
  \bibinfo{year}{2017}\natexlab{}.
\newblock \showarticletitle{Multiagent cooperation and competition with deep
  reinforcement learning}.
\newblock \bibinfo{journal}{{\em PloS one\/}} \bibinfo{volume}{12},
  \bibinfo{number}{4} (\bibinfo{year}{2017}), \bibinfo{pages}{e0172395}.
\newblock


\bibitem[\protect\citeauthoryear{Tieleman and Hinton}{Tieleman and
  Hinton}{2012}]%
        {tieleman2012lecture}
\bibfield{author}{\bibinfo{person}{Tijmen Tieleman} {and}
  \bibinfo{person}{Geoffrey Hinton}.} \bibinfo{year}{2012}\natexlab{}.
\newblock \showarticletitle{Lecture 6.5-rmsprop: Divide the gradient by a
  running average of its recent magnitude}.
\newblock \bibinfo{journal}{{\em COURSERA: Neural networks for machine
  learning\/}} \bibinfo{volume}{4}, \bibinfo{number}{2} (\bibinfo{year}{2012}),
  \bibinfo{pages}{26--31}.
\newblock


\bibitem[\protect\citeauthoryear{Wang and Yuan}{Wang and Yuan}{2015}]%
        {wang2015real}
\bibfield{author}{\bibinfo{person}{Jun Wang} {and} \bibinfo{person}{Shuai
  Yuan}.} \bibinfo{year}{2015}\natexlab{}.
\newblock \showarticletitle{Real-time bidding: A new frontier of computational
  advertising research}. In \bibinfo{booktitle}{{\em Proceedings of the Eighth
  ACM International Conference on Web Search and Data Mining}}. ACM,
  \bibinfo{pages}{415--416}.
\newblock


\bibitem[\protect\citeauthoryear{Wang, Liu, Liu, Hao, He, Hu, Yan, and Li}{Wang
  et~al\mbox{.}}{2017}]%
        {wang2017ladder}
\bibfield{author}{\bibinfo{person}{Yu Wang}, \bibinfo{person}{Jiayi Liu},
  \bibinfo{person}{Yuxiang Liu}, \bibinfo{person}{Jun Hao},
  \bibinfo{person}{Yang He}, \bibinfo{person}{Jinghe Hu},
  \bibinfo{person}{Weipeng Yan}, {and} \bibinfo{person}{Mantian Li}.}
  \bibinfo{year}{2017}\natexlab{}.
\newblock \showarticletitle{LADDER: A Human-Level Bidding Agent for Large-Scale
  Real-Time Online Auctions}.
\newblock \bibinfo{journal}{{\em arXiv preprint arXiv:1708.05565\/}}
  (\bibinfo{year}{2017}).
\newblock


\bibitem[\protect\citeauthoryear{Wu, Yeh, and Chen}{Wu et~al\mbox{.}}{2015}]%
        {wu2015predicting}
\bibfield{author}{\bibinfo{person}{Wush Chi-Hsuan Wu}, \bibinfo{person}{Mi-Yen
  Yeh}, {and} \bibinfo{person}{Ming-Syan Chen}.}
  \bibinfo{year}{2015}\natexlab{}.
\newblock \showarticletitle{Predicting winning price in real time bidding with
  censored data}. In \bibinfo{booktitle}{{\em Proceedings of the 21th ACM
  SIGKDD International Conference on Knowledge Discovery and Data Mining}}.
  ACM, \bibinfo{pages}{1305--1314}.
\newblock


\bibitem[\protect\citeauthoryear{Yuan, Wang, and Zhao}{Yuan
  et~al\mbox{.}}{2013}]%
        {yuan2013real}
\bibfield{author}{\bibinfo{person}{Shuai Yuan}, \bibinfo{person}{Jun Wang},
  {and} \bibinfo{person}{Xiaoxue Zhao}.} \bibinfo{year}{2013}\natexlab{}.
\newblock \showarticletitle{Real-time bidding for online advertising:
  measurement and analysis}. In \bibinfo{booktitle}{{\em Proceedings of the
  Seventh International Workshop on Data Mining for Online Advertising}}. ACM,
  \bibinfo{pages}{3}.
\newblock


\bibitem[\protect\citeauthoryear{Zhang, Yuan, and Wang}{Zhang
  et~al\mbox{.}}{2014}]%
        {zhang2014optimal}
\bibfield{author}{\bibinfo{person}{Weinan Zhang}, \bibinfo{person}{Shuai Yuan},
  {and} \bibinfo{person}{Jun Wang}.} \bibinfo{year}{2014}\natexlab{}.
\newblock \showarticletitle{Optimal real-time bidding for display advertising}.
  In \bibinfo{booktitle}{{\em Proceedings of the 20th ACM SIGKDD international
  conference on Knowledge discovery and data mining}}. ACM,
  \bibinfo{pages}{1077--1086}.
\newblock


\end{thebibliography}

\end{document}